\documentclass[onefignum,onetabnum]{siamart171218}

\usepackage{booktabs}

\usepackage{lipsum}
\usepackage{amsfonts}
\usepackage{graphicx}
\usepackage{epstopdf}
\usepackage{xr}
\ifpdf
  \DeclareGraphicsExtensions{.eps,.pdf,.png,.jpg}
\else
  \DeclareGraphicsExtensions{.eps}
\fi

\usepackage{listings, wrapfig, parcolumns}
\definecolor{codegreen}{rgb}{0,0.6,0}
\definecolor{codegray}{rgb}{0.5,0.5,0.5}
\definecolor{codepurple}{rgb}{0.58,0,0.82}
\definecolor{backcolour}{rgb}{0.95,0.95,0.95}

\lstdefinestyle{mystyle}{
	language=Python,
    backgroundcolor=\color{white},   
    commentstyle=\color{codegreen},
    keywordstyle=\color{black},
    numberstyle=\tiny\color{codegray},
    stringstyle=\color{codepurple},
    basicstyle=\ttfamily\small,
    breakatwhitespace=false,         
	breaklines=true,                 
	frame=single,
	columns=fullflexible,
    captionpos=b,                    
    keepspaces=true,                 
    numbers=right,  
    numberblanklines=true,
	showspaces=false,                
	texcl=true,
    showstringspaces=false,
	showtabs=false,                  
	keywordstyle = [2]{\color{magenta}},
    keywordstyle = [3]{\color{yellow}},
    keywordstyle = [4]{\color{teal}},
    otherkeywords = {Conv1D, Conv2D, BatchNormalization,AveragePooling3D,MaxPool2D,UpSampling3D, Dense, Conv3D,MaxPool1D},
    morekeywords = [2]{Conv1D, Conv2D, BatchNormalization,AveragePooling3D,MaxPool2D,UpSampling3D, Dense, Conv3D, MaxPool1D},
    morekeywords = [4]{import, lambda},
    tabsize=2
}
\def\cO{{\cal O}}

\def\vx{\mathbf{x}}
\def\vy{\mathbf{y}}
\def\vr{\mathbf{r}}
\def\vs{\mathbf{s}}

\def\RR{\mathbb{R}}

\usepackage{subfigure}

\usepackage{enumitem}

\newcommand{\rev}[1]{\textcolor{black}{#1}}

\newcommand{\wbnn}{\textsc{WideBNet}~}

\newsiamremark{remark}{Remark}
\newsiamremark{hypothesis}{Hypothesis}
\crefname{hypothesis}{Hypothesis}{Hypotheses}
\newsiamthm{claim}{Claim}

\headers{Multi-scale Butterfly Networks}{Li, Demanet, and Zepeda-N\'u\~nez}

\title{Wide-band butterfly network:  stable and efficient inversion via multi-frequency neural networks.\thanks{Submitted to the editors DATE}.
\funding{The authors thank Total SA for support. LD is also supported by AFOSR grant FA9550-17-1-0316. L.Z.-N. is supported in part by the Wisconsin Alumni Research Fund, the National Science Foundation under the grant DMS-2012292, and NSF TRIPODS award 1740707.}}

\author{Matthew Li\thanks{Computational Science and Engineering, Massachusetts Institute of Technology, Cambridge MA 02139
  (\email{mtcli@mit.edu})}
\and Laurent Demanet\thanks{Department of Mathematics and Earth Resources Lab, Massachusetts Institute of Technology, Cambridge MA 02139
  (\email{laurent@math.mit.edu}) }
\and Leonardo Zepeda-N\'u\~nez\thanks{Department of Mathematics, University of Wisconsin-Madison, Madison WI 53706
  (\email{zepedanunez@wisc.edu}) }}

\usepackage{amsopn}

\makeatletter
\newcommand*{\addFileDependency}[1]{
  \typeout{(#1)}
  \@addtofilelist{#1}
  \IfFileExists{#1}{}{\typeout{No file #1.}}
}
\makeatother

\ifpdf
\hypersetup{
  pdftitle={Wide-band Butterfly Networks},
  pdfauthor={M. Li, L. Demanet, and L. Zepeda-N\'u\~nez}
}
\fi

\begin{document}

\maketitle
  
\begin{abstract}

We introduce an end-to-end deep learning architecture called the \textit{wide-band butterfly network} (\textsc{WideBNet}) for approximating the inverse scattering map from wide-band scattering data. This architecture incorporates tools from computational harmonic analysis, such as the butterfly factorization, and traditional multi-scale methods, such as the Cooley-Tukey FFT algorithm, to drastically reduce the number of trainable parameters to match the inherent complexity of the problem. As a result, \wbnn is efficient: it requires fewer training points than off-the-shelf architectures, and has stable training dynamics which are compatible with standard weight initialization strategies. The architecture automatically adapts to the dimensions of the data with only a few hyper-parameters that the user must specify. \wbnn is able to produce images that are competitive with optimization-based approaches, but at a fraction of the cost, and we also demonstrate numerically that it learns to super-resolve scatterers with a full aperture configuration.

\end{abstract}



\section{Introduction}

There is nowadays extensive documentation on the remarkable ability of neural networks to approximate high-dimensional, non-linear maps provided that enough data are available~\cite{lecun2015deep}. In many applications the process of discovering such approximations simply involves enriching the network models, i.e., making them wider and/or deeper, until favourable stationary points arise in the empirical loss landscape. This practice can be partially justified by the asymptotic capacity of neural networks to approximate functions to within arbitrary accuracy, assuming only mild regularity conditions~\cite{CohenSharir2018,hornik90,Mhaskar2018}. Oftentimes, however, this strategy results in models that are vastly over-parametrized, even when compared to the already massive datasets that are necessary for training. For reasons that we outline below, these approximation-theoretic results also obscure many pre-asymptotic complications that are particularly acute when neural networks are applied to scientific applications. In these instances the neural architectures often require specific tailoring to the task at hand in order to satisfy the stricter requirements of scientific computing.

In this paper we focus on the problem of high-resolution imaging of scatterers arising from wave-based inverse problems. This task naturally arises in many scientific applications: e.g. biomedical imaging~\cite{Schomberg:1978}, synthetic aperture radar~\cite{Cheney_SAR:2001}, non-destructive testing~\cite{Pettit:2015}, and geophysics~\cite{Rawlinson:2010}. This problem also prototypically exhibits two challenges that are commonly encountered in scientific machine learning. First: obtaining the training data in this setting -- whether synthetically or experimentally -- comes at considerable expense, which bottlenecks the size of the models that can be reliably trained to satisfy the stringent accuracy requirements. This necessitates the use of unconventional architectures that are bespoke to each problem. Second: wave scattering involves \emph{non-smooth data} that are recordings of highly oscillatory, broadband, scattered waveforms. These highly oscillatory (i.e. high-frequency) signals are known to impede the training dynamics of many machine learning algorithms \cite{Fprinciple_NeuroIPS} and thus require new strategies to mitigate their effect.

Existing methods for scientific machine learning address the issue of data scarcity by ``incorporating underlying physics'' into the design of neural architectures.  In instances where the problem data are \emph{smooth}, this demonstrably reduces the total number of trainable weights which, in turn, reduces the number of training data required. Broadly categorized, these designs manifest as either: \emph{(i)} explicitly enforcing physical symmetries into the network \cite{DeepDensity,zhang2018deep,zhang2018deepcg}, \emph{(ii)} exploiting signal invariances and equivariances when processing the data \cite{Invariant_Scattering_CNN}, \emph{(iii)} directly embedding the governing differential equations into the objective function \cite{innes2019differentiable, Raissi2018}, or \emph{(iv)} imposing information flow (i.e., connectivity) within the architecture according to multi-scale interactions inherent to the physics of the data generating process \cite{MNNH2,Khoo_YingSwitchNet:2019}. Surprisingly, in addition to lowering data requirements these strategies are also observed to improve on the testing accuracy of comparable conventional models which are trained on a larger set of training points \cite{He2015ConvolutionalNN,Mhaskar2018,Zhang_al2020:learning_map}. 

In comparison, not much is known about designing architectures for processing \emph{non-smooth data} such as high-frequency waves. Here the same challenge that confounds the original inverse problem -- namely, the processing of highly oscillatory signals -- similarly obstructs direct application of machine learning methods. This idea is formalized by the ``F-principle''  conjecture \cite{Fprinciple_NeuroIPS} which documents the relation between machine learning methods and Fourier analysis. Specifically, it is empirically observed that models with fully-connected and convolutional architectures preferentially capture the low-frequency features of the target function. On the other hand, considerable expense (with respect to model size and/or data) is needed to learn high-frequency features \cite{Fprinciple_Theory}. Some examples even demonstrate that training can completely fail when the target function lacks low-frequency content even if highly expressive models are used \cite{MultiscaleDNN_high_dim_PDE,Fprinciple_Applications}. The F-principle thus demonstrates that although neural networks are universal approximators in an asymptotic sense, new strategies are needed to account for the issue with high frequencies if tractably computable models are to be obtained.

We note that in our application the forward and inverse maps are intrinsically oscillatory on account of the physics of wave propagation. This can be seen as an immediate consequence of the \emph{dispersion relation} in homogeneous media,
\begin{equation}
\label{eq:dispersion}
\lambda f = c,
\end{equation}
which describes the inverse scaling of the frequency $f$ of propagating waves to their spatial wavelength $\lambda$ by a factor of the local wave-speed $c$. This dispersion relation, in conjunction with rudimentary signal processing, effectively suggests that images generated by back-propagating the recorded waves into the medium are constrained to a wavelength dependent resolution limit, i.e., the classical diffraction limit \cite{garnier2016passive}. High resolution imaging of scatterers thus seemingly necessitates the use of high frequency waves to probe the media.

\subsection{Our Contributions}

We introduce a custom architecture for the inverse wave scattering problem which we call \textsc{WideBNet}. We demonstrate that our architecture overcomes the major deficiencies outlined above for traditional architectures. Specifically, \wbnn relies on ideas from the butterfly factorization~\cite{Li_Yang_Martin_Ho_Ying:Butterfly_Factorization} to capture the Fourier Integral Operators (FIOs) underlying the physics of wave-scattering -- as a result, fewer training datapoints are needed. Moreover, it addresses the high frequency limitations identified by F-principle by mimicking the Cooley-Tukey algorithm~\cite{Cooley_Tukey:1965} to process multi-frequency data only at localized length scales -- this effectively renders each frequency slice as \emph{locally} low-frequency information. These design choices afford \wbnn the following benefits compared to off-the-shelf deep learning models:

\textbf{ Training Efficiency } 
The architecture builds upon the butterfly factorization and thus systematically adapts to the input size of the data, i.e., the number of pixels in the image. As a result, the degrees of freedom in the model scale near-linearly with the input size, and the depth of the network scales logarithmically with the input size\footnote{When compared to other machine learning based approaches, we note that a comparable implementation using fully connected networks results in models with degrees of freedom that scale cubically with the size of the input, i.e., the number of pixels in the image, and are thus prohibitively expensive to train. Conversely, a purely convolutional neural network implementation for the task requires far deeper networks (or far wider filters) to properly capture the long-range interactions governed by the underlying wave physics. Such deep networks are known to exhibit issues with exploding/vanishing gradients leading to unstable training dynamics \cite{Bengio1994}. While we do not discount the possibility of other hybridized (fully connected + convolutional) architectures which achieve the same task, we emphasize that these architectures would not be immediately transferable for different image and data resolution requirements.}. This makes training our network data-efficient as there are relatively fewer degrees of freedom. 

\textbf{ Training Stability } \wbnn avoids empirically observed shortcomings with other network architectures that rely on the butterfly factorization. For example, in \cite{Yingzhou2018} the authors prove that butterfly-networks are capable of efficiently approximating generic FIOs, but report that learning such operators requires an accurate initialization to avoid local minima; this is typically not easily obtainable for most FIOs, including our application. Similarly, \cite{Khoo_YingSwitchNet:2019} introduces a butterfly-network for single frequency inversion but requires increasing the width of their network (so that the degrees of freedom no longer scale linearly) to overcome local minima. In contrast, empirically we observe that \wbnn does not require specialized initialization strategies, it does not routinely get stuck in local minima, and it does not exhibit exploding or vanishing gradients. We speculate that the training stability of \wbnn can be attributed to its use of multi-frequency data that are banded to appropriate length scales to avoid the F-principle limitations. 

\textbf{Imaging Super-resolution}  In our numerical results we demonstrate that our network super-resolves scatterers, i.e., produces sharp images of sub-wavelength features\footnote{We plan to further investigate and document this super-resolution phenomenon in forthcoming work.} such as diffraction corners, in addition to producing competitive images when compared against classical optimization-based inversion methods in the traditional super-diffraction regime.

\textbf{ Hyper-parameter Efficiency } 
It is efficient to tune the hyper-parameters of \wbnn as there are only a few which are used to describe the architecture. We note that in numerical examples we observe strong robustness to variations in these hyper-parameters. This indicates that relatively little effort is needed on the user's part to optimally tune our architecture.

A detailed discussion of the \wbnn architecture, as well as implementation notes, can be found in Section~\ref{sec:architecture}. Meanwhile, we briefly sketch the intuition behind the design choice here. The idea to embed the butterfly factorization into the architecture is to effectively furnish our network with a strong prior on the physics of wave scattering. Indeed, we provide numerical evidence that it is necessary to manually encode the long range ``non-local'' interactions between scatterers and sources that are inherent to the wave kernel. Mathematically these interactions are known to described as the action of an FIO \cite{Hormander:The_Analysis_of_Linear_Partial_Differential_Operators_FIO}, which can be discretely represented in a complexity-optimal manner by means of the butterfly factorization \cite{Li_Yang_Martin_Ho_Ying:Butterfly_Factorization} and the butterfly algorithm \cite{Candes_Demanet_Ying:A_Fast_Butterfly_Algorithm_for_the_Computation_of_Fourier_Integral_Operators,ONeil_2010,Borm_Butterfly:2017}. 

However we stress that the marriage of the butterfly factorization with network architectures is \emph{not} the original contribution of this work; butterfly-like architectures  have  been  previously proposed by other authors, albeit with  different goals \cite{Yingzhou2018,Khoo_YingSwitchNet:2019}, and we review these contributions below in Section~\ref{sec:references}. Instead, our contribution is the \emph{combination of this network architecture with multi-frequency data}. This data assimilation strategy takes cues from the Cooley-Tukey algorithm and is done, in part, to address the F-principle. For reference, one notable strategy for avoiding the F-principle involves partitioning the model into disjoint Fourier segments and frequency down-shifting accordingly \cite{cai2019phasednn}, but this introduces costly convolutions in data-space and requires a dense data sampling strategy that scales unfavourably with dimensionality. Our network improves on this approach by exploiting the duality between frequency $f$ and wavelength $\lambda$, as described by the dispersion relation in \eqref{eq:dispersion}, to introduce data only at their local length scales. This effectively performs frequency downshifting by \emph{spatial downsampling}. This strategy is easily accommodated by the butterfly architecture as these multi-scale interactions are already implicitly present in its formulation.  

\section*{Outline}
The remainder of this document is structured as follows. We close this section with relevant background material on existing algorithms for inverse scattering and relevant machine learning based approaches for general inverse problems in Section \ref{sec:references}. Section \ref{sec:background} describes the technical details of the underlying physical model, provides background on the problem to solve and the algorithmic ideas behind the network. In Section \ref{sec:architecture} we present in detail the network architecture. Finally, in Section~\ref{sec:numerical_results} we present and discuss the numerical results.

\subsection{Related Literature} \label{sec:references}

\subsubsection{Classical Approaches}
One of the earliest modalities in imaging is travel-time tomography \cite{Oldham1906,Gutenberg1914,Backus_Gilbert:1968}, in which the travel time of a wave passing between two points is used to reconstruct the medium wave-speed \cite{Stefanov2019}. 
Travel-time tomography is a rather mature technique, which can even be easily and cost effectively implemented in portable ultra-sound devices \cite{ultra_sound_2015}. However, its resolution deteriorates greatly when dealing with highly heterogeneous media and in the presence of multiple scattering.

In response to these drawbacks, several techniques were developed such as reverse time migration \cite{RTM:1983}, linear sampling method \cite{Colton_Kirsh:LSL1996}, decomposition methods \cite{Kirsh:Factorization_methods} among many others. See \cite{Colton_Kress:Inverse_Acoustic_and_Electromagnetic_Scattering_Theory} and \cite{Virieux_Operto:An_overview_of_full-waveform_inversion_in_exploration_Geophysics} for excellent historical reviews. 

Finally, a high-resolution technique, called full-waveform inversion (FWI) \cite{Taratola:Inversion_of_seismic_reflection_data_in_the_acoustic_approximation} was developed in the late 80s, which has been shown empirically capable of handling multiple scattering. FWI solves a constrained optimization problem in which the misfit between the real data and synthetic data coming from the numerical solution of the PDE is minimized. This technique, coupled with large computing power, has been successful at recovering the properties of the sub-surface \cite{Pratt:Seismic_waveform_inversion_in_the_frequency_domain;_Part_1_Theory_and_verification_in_a_physical_scale_model}. Nowadays, it is considered the gold standard in geophysical exploration \cite{Virieux_FWI:2017}.  

Despite its enormous success, FWI still suffers from three significant challenges: prohibitive computational cost, cycle-skipping and limited resolution. The prohibitive computational cost is linked to the cost of computing the gradient within the optimization loop, which requires a large amount of wave solves. The resulting complexity of each iteration is quadratic\footnote{Using state-of-the-art sparse direct solvers. It can be further reduced to $\cO(N^{3/2})$ using state-of-the-art pre-conditioner, but with substantially larger constants.} \cite{Borges_Gillman_Greengard:2017} with respect to number of unknowns to recover. Progress in this direction has focused on developing fast PDE solvers \cite{ZepedaDemanet:the_method_of_polarized_traces,EngquistYing:Sweeping_PML} which are necessary to compute the gradient. In addition, numerous iterations are usually required for convergence. This prohibitive computational cost has hampered the application of this vastly superior technique to domains where images are required on-the-fly, such as biomedical imaging.

 Cycle-skipping refers to the undesirable convergence to spurious local minima by the FWI algorithm. This effect is especially pronounced when low-frequency data are scarce as these determine the kinematically relevant, low-wavenumber components of the material properties. Unfortunately, acquiring low-frequency data from practical field applications is a challenging and expensive task. As such, research in this area has focused on regularizing the optimization objective to handle the lack of low-frequency data \cite{Symes_Carazzone:1991,Leeuwen_Herrmann:2013}, using a smooth initial guess from travel-time tomography \cite{Alkhalifah:2014}, or extrapolating the low-frequency component from higher frequency data \cite{Li_Demanet:2016}. Lastly, quantifying the resolution limits of FWI remains an open problem \cite{Fichtner2011}, i.e., understanding the finest details available by the algorithm and its scaling with respect to the shortest wavelength at which data are available. This is important for applications requiring accurate images of  discontinuities \cite{Atkinson:1997,burger_osher_2005,de_Buhan:2013,deBuhan:2017}, such as those arising in natural geophysical formations, for properly detecting cracks and dislocations in materials, or for detecting and interpreting anomalies in biomedical imaging. 


\subsection{Machine Learning Approaches}

Besides the classical, PDE constrained optimization approaches, several recent methodologies based on machine learning for more general inverse problems have been proposed lately. 

In \cite{PINN_Inverse_Problems} authors used the recently introduced paradigm of physics informed neural networks (PINN) to solve for inverse problems in optics. Aggarwal et al.~introduce a model-based image reconstruction framework \cite{MoDL} for MRI reconstruction. The formulation contains a novel data-consistency step that performs conjugate gradient iterations inside the unrolled algorithm. Gilton et al.~proposed in \cite{Neumann_Networks} a novel network based on Neumann series coupled with a hand-crafted pre-conditioner for linear inverse problems, which recast an unrolled algorithm as elements of a Neumann series. In \cite{Mao:2016} Mao et al.~use a deep encoder-decoder network reminiscent of U-nets \cite{U-Net} for image de-noising, using symmetric skip connections. 

In \cite{FanYing:scattering} the authors proposes a rotationally equivariant network for inverse scattering, that is only valid for homogeneous media; the same type of ideas is applied to travel-time tomography \cite{FanYing:traveltime} and optical tomography \cite{FanYing:RTE}.

Among the more general field of computational harmonic analysis, to which the butterfly algorithm is connected, we mention several other applications.  Networks based on the Short-time Fourier transform \cite{STFT-Nets,DeepSense2017} has been used for hierarchically decomposing signals in a non-linear fashion. 
Networks based on the scattering transform has been proposed \cite{Invariant_Scattering_CNN} to take in account translation invariance in images. In \cite{Framelets} the authors introduced another framework based on frames for inverse problems, which was applied to computer tomography de-noising \cite{Framelets_denosing}.

In addition, machine learning recently has been used for super resolution in the signal processing context \cite{Candes_Fernandez_Granda:Towards_a_Mathematical_Theory_of_Super_resolution} and image processing. Recently newly developed frameworks such as generative adversarial networks (GANs) \cite{goodfellow2016dl,GANs:2014}, and variational autoencoders (VAEs) \cite{Kingma2014a,Rezende_Mohamed_Wierstra_2014} have been used for super resolution in the context of image processing \cite{Isolapix2pix:2016,Ledig2017,mirzacGANs:2014}. These techniques provide an end-to-end map that relies on the statistical properties of the images to super-resolve them. 

\rev{Another related approach is the recently introduced Fourier Neural Operators~\cite{FourierNeuralOp} that aims to learn the Fourier multipliers in a context akin to pseudo-differential operators using an aggressive filtering, which is compensated by the non-linear activation functions. Although this approach captures long-range interactions, it is unclear whether the highly oscillatory behavior of wave data can be captured efficiently.}

The method introduced in this manuscript follows similar ideas to \cite{MNNH2,linear_butterfly}, where the authors introduce tools from numerical analysis into deep learning. They build on the sparse matrix factorizations that result from exploiting low-rank interactions arising from the underlying physics of the problem. These factorizations are translated into the machine learning context: each matrix factor becomes a layer in the network wherein the sparsity pattern informs the connectivity between layers, and the matrix entries themselves are viewed as learnable weights. In particular, the authors translate hierarchical matrices ($\mathcal{H}$-matrices), which are factorizations of operators into low-rank and permutations matrices, into individual layers in neural network architectures. Although these networks are well suited for smooth data with compressible long range interactions, which is the underlying motivation for the $\mathcal{H}$-matrices, they are not well suited for wave-scattering problems where the data are highly oscillatory, and where the long-range interactions are not typically compressible. 

Instead, the correct idea for capturing wave propagation is the choice of the butterfly factorization, as motivated by their use for representing FIOs. In fact, architectures based on butterfly algorithm have been previously proposed, albeit with different goals as the one considered in this paper. In \cite{linear_butterfly} the authors recover the butterfly structure of certain linear operators, from permutation operations. In \cite{Khoo_YingSwitchNet:2019} the authors use a one-level butterfly network with applications to inverse scattering, though critically they require a super-linear scaling in their number of parameters. In \cite{Yingzhou2018} the authors propose a mono-chromatic butterfly network similar to the architecture used in this case, which was later simplified in \cite{Butterfly-Net2}. In \cite{linear_butterfly}, the authors use the backbone of the butterfly structure to learn fast matrix approximations, with a clever variational relaxation strategy for learning the permutation factors. However, as mentioned in the prequel, none of these works address the use of butterfly factorizations for super-resolution in wave-based imaging which requires stable training over a wideband dataset. 

\section{Background}  \label{sec:background}
In this section we briefly review concepts from classical imaging (see \cite{Colton_Kress:Integral_Equation_Methods_in_Scattering_Theory} for further details) and their connection with fast numerical methods. We also provide a succinct description of the butterfly factorization and Cooley-Tukey FFT algorithm to motivate the discussion of our architecture in Section~\ref{sec:architecture}.

\subsection{Underlying Physical Model}  \label{sec:phys_model}

\begin{figure}
    \centering
    \includegraphics[width=0.5\textwidth]{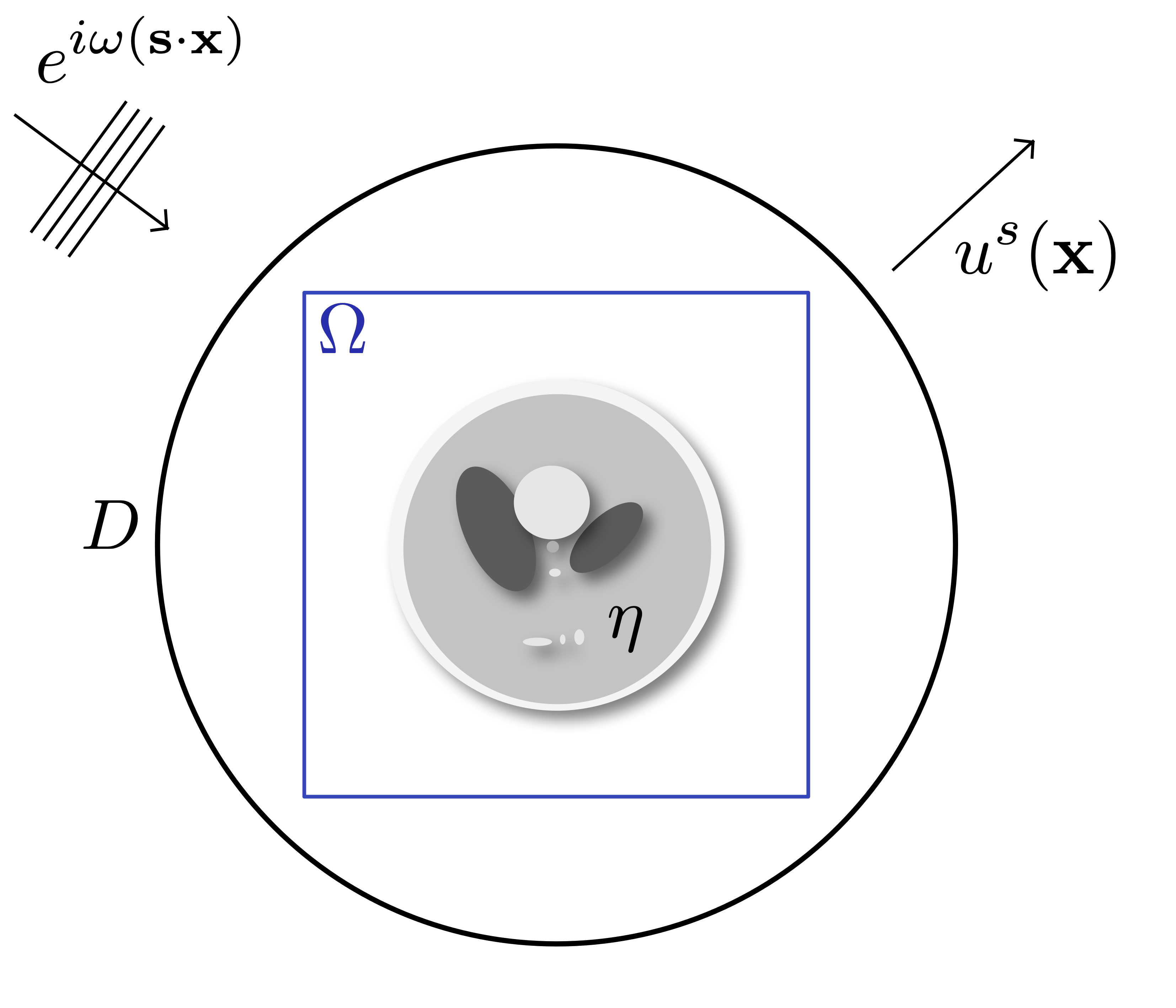}
    \caption{Diagram of the inverse scattering problem. We probe the medium with a plane-wave with direction $\vs$, and we sample the scattered field on the disk $D$.}
    \label{fig:scattering}\textbf{}
\end{figure}

We consider the time-harmonic wave equation with constant-density acoustic physics, also called the Helmholtz equation, with frequency $\omega$ and squared slowness $m$, given by  
\begin{equation} \label{eq:helmholtz}
    (\Delta + \omega^2 m(\vx)) u (\vx) = 0
\end{equation}
with radiating boundary conditions. We further suppose the slowness squared admits a scale separation into
\begin{equation*}
    m(\vx) = m_0(\vx) + \eta(\vx),
\end{equation*}
where $m_0$ corresponds to the smooth background slowness, assumed known, and $\eta$ the rough perturbation that we wish to recover. If the background slowness is constant and normalized\footnote{This assumption is only made to make the presentation more transparent.} so that 
\begin{equation*}
    m(\vx) = 1 + \eta(\vx),
\end{equation*}
then solutions to \eqref{eq:helmholtz} can be expressed in the form
\begin{equation} \label{eq:wavefield_decomposition}
    u(\vx) = e^{i\omega \left ( \mathbf{s} \cdot \vx \right )} + u^{sc}(\vx), 
\end{equation}
where $e^{i\omega \left ( \mathbf{s} \cdot \vx \right ) }$ is the incoming plane wave, with propagating direction $\mathbf{s}$, that we use to ``probe'' the perturbation, and $u^{sc}(\vx)$ is the scattered field produced by the interaction of the perturbation with the impinging wave. The scattered field satisfies
\begin{equation} \label{eq:scattering}
    \left \{   \begin{array}{ll}
                \left(\Delta+\omega^{2}( 1 + \eta(\vx)\right) u^{sc}(\vx) = -\omega^2 \eta(\vx) e^{i \omega ( \mathbf{s} \cdot \vx)} & \text{ for } \vx \in \mathbb{R}^2,  \\ \displaystyle
                \lim _{|\vx| \rightarrow \infty}|\vx|^{1/2}\left(\frac{\partial}{\partial|\vx|}-\mathrm{i} \omega\right) u^{\mathrm{s}}(\vx)=0,
                \end{array}
    \right .
\end{equation} 
following the configuration depicted in Fig.~\ref{fig:scattering}.
We select the detector manifold~$D$ to be a circle of radius~$R$ that encloses the domain of interest~$\Omega$. For each incoming direction $\mathbf{s} \in \mathbb{S}^1$, as defined in \eqref{eq:scattering}, the data are given by sampling the scattered field with receiver elements that are located on $D$ and indexed by $\mathbf{r} \in \mathbb{S}^1$. We assemble the data for each frequency~$\omega$ into a matrix~$\Lambda^\omega$ whose $(\vs,\vr)$-th entry corresponds to
\begin{equation} \label{eq:far_field_pattern_def}
    \Lambda_{\vs, \vr} = u^{sc}(R\vr;\vs),
\end{equation}
where we omit the dependence on $\omega$ in the right hand side. We call $\mathcal{F}^{\omega}[\eta]$ the {\it forward map} relating the perturbation $\eta$ to the data matrix~$\Lambda^\omega$\footnote{\rev{We point out that the data is not linearized, we solve \eqref{eq:scattering}, which depends non-linearly on $\eta$, to obtain the scattered wavefield, $u^{sc}$, for each incoming direction. One can easily recover the full wavefield using \eqref{eq:wavefield_decomposition}.}}.

Accordingly, we can cast the inverse problem for recovering the rough perturbation as
\begin{equation} \label{eq:fwi}
    \eta^* = \text{argmin}_{\mu} \|\mathcal{F}^{\omega}[\mu] - \Lambda^\omega \|_2^2.
\end{equation}
Linearizing the forward operator $\mathcal{F}^{\omega}$ is instructive as it sheds light on the essential difficulties of this problem.  Using the classical Born approximation in \eqref{eq:scattering} we obtain that 
\begin{equation}
    u^{sc}(\mathbf{x}) = \omega^2 \int_{\RR^2} \Phi^{\omega}(\mathbf{x}, \mathbf{y})  \eta(\mathbf{y}) e^{i \omega ( \mathbf{s} \cdot \mathbf{y})} d\vy,
\end{equation}
where  $\Phi^{\omega}$ is the Green's function of the two-dimensional Helmholtz equation in homogeneous media, i.e.,  $\Phi^{\omega}$ satisfies
\begin{equation} \label{eq:GreenFunction}
    \left \{   \begin{array}{ll} \displaystyle
                \left(\Delta+\omega^{2}\right) \Phi^{\omega}(\vx, \vy) = -\delta(\vx, \vy) & \text{ for } \vx \in \mathbb{R}^2,  \\ \displaystyle
                \lim _{|\vx| \rightarrow \infty}|\vx|^{1/2}\left(\frac{\partial}{\partial|\vx|}-\mathrm{i} \omega\right) \Phi^{\omega}(\vx, \vy) =0.
                \end{array}
    \right .
\end{equation} 
Furthermore, we can use the classical far-field asymptotics of the Green's function to express
\begin{equation}
    u^{sc}(R \vr) = -\omega^2 \frac{e^{i \omega R}}{\sqrt{R}} \int_{\RR^2}  \eta(y) e^{i \omega ( \mathbf{s} - \vr) \cdot \mathbf{y})} d\vy + \mathcal{O}(R^{-3/2}).
\end{equation}
Thus, up to a re-scaling and a phase change, the far-field pattern defined in \eqref{eq:far_field_pattern_def} can be approximately written as a Fourier transform of the perturbation, viz., 
\begin{equation} \label{eq:far_field_pattern}
    \Lambda_{\vs, \vr}(\omega) \approx F^{\omega} \eta = -\omega^2 \frac{e^{i \omega R}}{\sqrt{R}} \int_{\RR^2}  e^{i \omega ( \mathbf{s} - \vr) \cdot \mathbf{y}} \eta(y) d\vy. 
\end{equation}
In this notation $F^\omega$ is the linearized forward operator acting on the perturbation. 

Solving the inverse problem \eqref{eq:fwi} using the linearized operator in \eqref{eq:far_field_pattern} and Tikhonov-regularization with regularization parameter~$\epsilon$ results in the explicit solution
\begin{equation}\label{eq:FBP}
    \eta^* = \left ( \left(F^{\omega} \right )^* F^{\omega} + \epsilon I \right )^{-1} \left(F^{\omega} \right )^*  \Lambda^\omega.
\end{equation}
This formula is also referred to as filtered back-projection \cite{Colton_Kress:Integral_Equation_Methods_in_Scattering_Theory}, is optimal with respect to the $L^2$-objective and, concomitantly, tends to yield low-pass filtered estimates, particularly with large~$\epsilon$. In practice~$\epsilon$ is chosen to be sufficient large so as to remedy the ill-conditioning of the normal operator $\left(F^{\omega} \right )^* F^{\omega}$.

Performing the inversion numerically requires discretizing the wavespeed and the sampling geometry. We discretize $\Omega$ using $N = n_x \times n_z$ degrees of freedom following the Nyquist sampling rate of  $n_x \sim n_z \sim \omega$. The scattered data $\Lambda^\omega$ are discretized into an $n_{src} \times n_{rcv}$ matrix.


\begin{figure}[htp!]
     \centering
     \includegraphics[width=\textwidth,trim={00mm 0mm 00mm 00mm},clip]{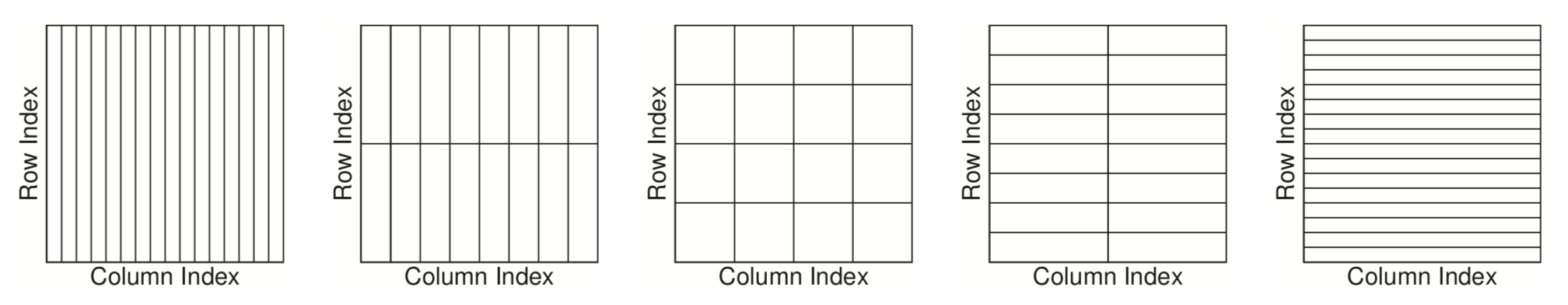}
     \caption{Sketch of a matrix exhibiting a complementary low-rank property. Each of the blocks induced by the different partitions has the same $\epsilon$-rank.}
     \label{fig:low_rank_butterfly}
 \end{figure} 

After discretization and a change of variables, $(F^{\omega})^*$ in \eqref{eq:FBP} is a Fourier transform (which itself is an FIO), and $ F^*F$ is a pseudo-differential operator, which, \rev{when the background medium is constant, is translation invariant, thus $\left (F^* F  + \epsilon I \right )^{-1}$ can be reduced to a convolution-type operator.} \rev{In more general situations of smooth background media the operator $\left (F^* F  + \epsilon I \right )^{-1}$ can be approximated by networks specifically tailored for pseudo-differential operators, such as the multiscale-neural network \cite{MNNH2}.}

\textbf{Remark:} Thus far we have assumed that we probe the perturbation $\eta$ using only a monochromatic time-harmonic wave with fixed frequency~$\omega$. As mentioned in the introduction this is known to be ill-posed and data at additional frequencies are required to stabilize the reconstruction \cite{Hahner_Hohage2001_inverse_problem_estimates}. In particular, a time-domain formulation known as the \emph{imaging condition} yields a more stable reconstruction using the full frequency bandwidth; this formula can be formally stated as 
\begin{equation}
\eta^* = \int_{\mathbb{R}} \left ( \left(F^{\omega} \right )^* F^{\omega} + \epsilon I \right )^{-1} \left(F^{\omega} \right )^*  \Lambda^\omega d \alpha(\omega),
\end{equation}
where $d \alpha(\omega)$ is a density related to the frequency content of the probing wavelet. When the density is well approximated by a discrete measure then 
\begin{equation} \label{eq:imaging_cond}
\eta^* \approx \sum_{i=1}^{N_{\text{freqs}}} \left ( \left(F^{\omega_i} \right )^* F^{\omega_i} + \epsilon I  \right )^{-1} \left(F^{\omega_i} \right )^*  \Lambda^{\omega_i} \alpha(\omega_i),
\end{equation}
over a discrete set of frequencies $\{ \omega_i \}_{i=1}^{N_{\text{freqs}}}$. We note that the selection of these frequencies, in addition to the optimal ordering in which the summation is computed under an iterative regime, remains an open question and an area of active research \cite{Borges_Gillman_Greengard:2017}. 

\subsection{Butterfly Factorization and Fourier Integral Operator}
\begin{figure}[ht]
     \centering
     \includegraphics[width=\textwidth,trim={00mm 0mm 0mm 00mm},clip]{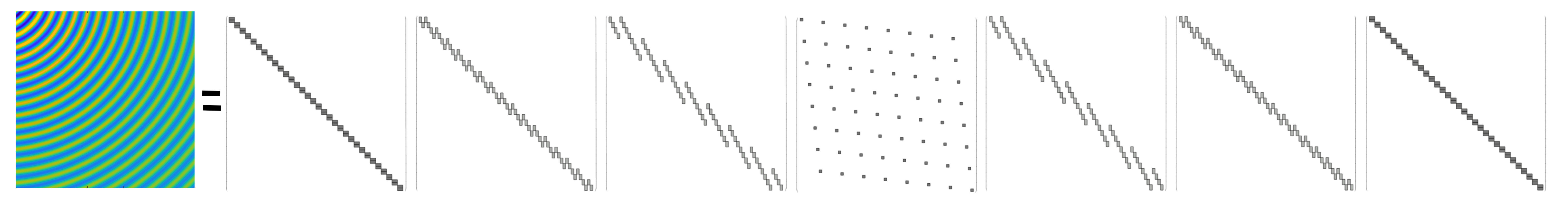}
     \caption{Sketch of the butterfly factorization, where the matrix at the left is factorized in sequence of very sparse matrices with a distinct sparsity pattern, induced by Fig.~\ref{fig:low_rank_butterfly}.}
     \label{fig:middle_factorization}
 \end{figure}

When the scattered field is given by \eqref{eq:far_field_pattern} then one could apply the fast Fourier transform \cite{Cooley_Tukey:1965} to compute the estimate \eqref{eq:imaging_cond} in quasi-linear time. However, with a heterogeneous background the linearized forward map is instead given by a more general representation usually known as a Fourier integral operator (FIO), which has the form 
\begin{equation} \label{eq:Fourier_operator}
(F^{\omega} \eta)(\vx)=\int_{\mathbb{R}^{2}} a_{\omega}(\vx, \vy) e^{i \omega \phi(\vx, \vy)}  \eta(\vy) d \vy.
\end{equation}
Here $\phi(\vx, \vy)$ is referred to as the phase (or travel-time) function while \rev{$a_{\omega}$} is typically a very smooth function that encodes the amplitude\footnote{The principal symbol $a_{\omega}$ \rev{depends asymptotically on $\omega$ as $\mathcal{O}(\omega^{-1})$} \cite{FIO:elastic_wave_propagation}.}. The work of \cite{Candes_Demanet_Ying:A_Fast_Butterfly_Algorithm_for_the_Computation_of_Fourier_Integral_Operators,Poulson_Demanet:a_parallel_butterfly_algorithm,ONeil_2010} recognized that even in this more generalized instance the application of $F^{\omega}$ and its adjoint can be computed in optimal complexity by means of the butterfly algorithm. The butterfly algorithm is a multi-scale algorithm which takes advantage of the \textit{complementary low-rank property} of the discretized operator depicted in Fig.~\ref{fig:low_rank_butterfly}. In its original form the algorithm relies on explicit knowledge of the phase function; later, in \cite{Li_Yang_Martin_Ho_Ying:Butterfly_Factorization} the authors introduced the butterfly factorization, which approximates the discretized operator \eqref{eq:Fourier_operator} by the multiplication of sparse matrices with a \emph{specific} sparsity pattern\footnote{This pattern is for the one-dimensional butterfly factorization, which already captures the key algorithmic ideas while keeping the presentation clean of ordering issues that arises in higher dimension.} as shown in Fig.~\ref{fig:middle_factorization}. 
 
In a nutshell, the butterfly factorization approximately factorizes a matrix $A$ that satisfies the complementary low-rank property in $L+3$ sparse matrices following: 
\begin{equation} \label{eq:butterfly_fact}
     A \approx A_{butterfly} = U^{L} G^{L-1} \cdots G^{L/2} S^{L/2}\left(H^{L/2}\right)^{*} \cdots\left(H^{L-1}\right)^{*}\left(V^{L}\right)^{*},
\end{equation}
where $U^L$ and $V^L$ are block diagonal matrices, $S^{L/2}$ is a weighted permutation matrix, usually called a \emph{switch matrix}, and $L$ is the number of levels in the factorization, which is usually a power of two. 

We can interpret the factors in \eqref{eq:butterfly_fact} following the original butterfly algorithm. $V^L$ extracts a local representation of the vector, then each factor $H^{\ell}$ compresses two neighboring local representations, i.e., decimates by a factor of two the number of local representations, while increasing the amount of information in each presentation. The switch matrix $S^{L/2}$ quickly redistributes the information contained in each local representation. The factors $G^{\ell}$ decompress the information contained in each representation at each stage, i.e., the local representations are split in two by each factor increasing the spatial resolution, and finally the factor $U^{L}$, transforms the local representations to the sampling points. 

For the sake of completeness we provide a formal argument to show that the FIO in \eqref{eq:Fourier_operator} satisfies the complementary rank property (see \cite{Candes_Demanet_Ying:A_Fast_Butterfly_Algorithm_for_the_Computation_of_Fourier_Integral_Operators} for a more comprehensive argument). \rev{In a nutshell, the complementary rank property for a matrix is the property in which each block of the partition in Fig.~\ref{fig:low_rank_butterfly} have $\epsilon$-ranks bounded by the same constant. Equivalently,  any block in which the multiplication of its sides is equal to $\mathcal{O}(N)$ has a bounded $\epsilon$-rank.}

Suppose that we have two points $\vx_0$ and $\vy_0$ in the evaluation and integration region respectively. We define two neighborhoods around each point, such that $|\vx-\vx_0| < d_x$ and $|\vy - \vy_0|<d_y$. \rev{In this case, $d_x$ and $d_y$ are the sides of the blocks, in physical space, shown in Fig.~\ref{fig:low_rank_butterfly}.} We then seek to find the largest values of $d_x$ and $d_y$ such that we can efficiently approximate 
\begin{equation}
\int_{|\vy - \vy_0|<d_y} a(\vx, \vy) e^{i \omega \phi(\vx, \vy)} \eta(\vy) d \vy,
\end{equation}
using a separable function. The principal symbol, $a(\vx, \vy)$ is supposed to be smooth and independent of $\omega$ (or weakly dependent), so we can focus our discussion to the oscillatory term $e^{i \omega \phi(\vx, \vy)}$.
 
 Using a Taylor expansion we have that 
 \begin{align*}
     \phi(\vx, \vy) = & \,  \phi(\vx_0, \vy_0) + \partial_{\vx} \phi(\vx_0, \vy_0) \cdot (\vx-\vx_0) + \partial_{\vy} \phi(\vx_0, \vy_0) \cdot (\vy-\vy_0) \\
                    + &  (\vx-\vx_0)^T \cdot \partial^2_{\vx} \phi(\vx_0, \vy_0) \cdot (\vx-\vx_0) +
                     (\vy-\vy_0)^T \cdot \partial^2_{\vy} \phi(\vy_0, \vy_0) \cdot (\vy-\vy_0) \\
                     + & 2 (\vx-\vx_0)^T \cdot \partial^2_{\vx, \vy} \phi(\vx_0, \vy_0) \cdot (\vy-\vy_0) + \mathcal{O}(d_x d_y)
 \end{align*}
Clearly the first five terms provide separable expressions, the sixth term can be easily bounded producing 
\begin{equation}
    e^{i\omega \phi(x,y)} =  e^{i\omega \psi(x)}e^{i\omega \xi(y)} ( 1 + \mathcal{O}(\omega d_x d_y))
\end{equation}
thus as long as $d_x d_y \leq \omega^{-1}$, then $e^{i\omega \phi(x,y)}$ can be locally approximated by a separable function. In the discrete case this property is translated to the fact that the multiplication of the height and the width of each block has a constant $\epsilon$-rank, which is exactly the complementary low-rank property showcased in Fig.~\ref{fig:low_rank_butterfly}.

\textbf{Remark:} We point out that there exist three different types of butterfly factorizations. The left one-sided, the right one-sided, and the two-sided (see \cite{Butterfly-factorization:Liu_Xing2020} for a review). In this work we focus on the two-sided version, which provides the best complexity. It is possible to ``neuralize'' the other two types of factorizations, which yield a specific type of CNN networks with sparse channel connections as shown in \cite{Butterfly-Net2}. 

\subsection{Cooley-Tukey Algorithm} \label{sec:FFT_algorithm}

The Cooley-Tukey FFT algorithm \cite{Cooley_Tukey:1965} is one of the most important algorithms in the 20th century \cite{Top10Algorithms}. It aims to compute the discrete Fourier transform (DFT) of a signal $\{x_{n}\}_{n=0}^{N-1}$ given by 
\begin{equation} \label{DFT}
    \hat{x}(k) = \sum_{k=0}^{N-1} x_n e^{-\frac{2\pi i}{N} n k},
\end{equation}
in $N \log{N}$ time. The algorithm leverages the algebraic structure of the N-th complex roots of the unit to recursively split the computation. The simplest version of the algorithm is called the radix-2 decimation-in-time FFT, which computes the DFT of both even-indexed and odd-indexed inputs, which are then merged to produce the final result. In particular, for the first level the DFT is rearranged as 
\begin{align*}
     \hat{x}(k) =& \sum_{m=0}^{N / 2-1} x_{2 m} e^{-\frac{2 \pi i}{N / 2} m k}+e^{-\frac{2 \pi i}{N} k} \sum_{m=0}^{N / 2-1} x_{2 m+1} e^{-\frac{2 \pi i}{N / 2} m k}, \\
         = & \,\,  \hat{x}_{e}(k) +e^{-\frac{2 \pi i}{N} k} \hat{x}_{o}(k),
\end{align*}
where $\hat{x}_{e}(k)$ and $\hat{x}_{o}(k)$ stand for the even and odd downsampled DFTs respectively. However, given that we are using decimated DFTs this expression is only valid for $k=0,... N/2-1$. Thus, in order to obtain the full length DFT, one can use the periodicity of the complex exponential, and we have that
\begin{align}
    \hat{x}(k)      = & \,\,\hat{x}_{e}(k) + e^{-\frac{2 \pi i}{N} k} \hat{x}_{o}(k), \label{eq:FFT_even}\\
     \hat{x}(k+N/2) = & \,\,  \hat{x}_{e}(k) - e^{-\frac{2 \pi i}{N} k} \hat{x}_{o}(k). \label{eq:FFT_odd}
\end{align}



\subsection{Wide-Band Butterfly Algorithm}
For the sake of simplicity we motivate the idea behind this paper, which is the multi-scale decomposition of the butterfly factorization, by using the Cooley-Tukey FFT algorithm. We point out that the same argument can be obtained from a rather involved analysis of the original butterfly algorithm. In particular, one can follow the description of the algorithm in \cite{DemanetYing:FIO} to show that if we build a compressed FIO, as the one in \eqref{eq:Fourier_operator}, at frequency $\omega$ using the butterfly algorithm, then most of the computation can be reused to build the same FIO, but at frequency $\omega/2$. 

The cornerstone of the approach is to leverage the recursive nature of the FFT algorithm to reuse most of the algorithm pipeline when computing the FFT of decimated signals, or in the case of \eqref{eq:Fourier_operator} at lower frequencies. We focus our attention on two operations: computing the DFT of a decimated signal using the FFT for a non-decimated signal, computing the same DFT using a decimated algorithm, but keeping a non-decimated resolution. These two operations will be key when designing our network.

From \eqref{eq:FFT_even} and \eqref{eq:FFT_odd} we clearly see that we can compute the DFT of a decimated signal, using the regular FFT algorithm. One only needs to interweave the original signal with zeros, then apply the FFT for the longer signal, and then truncate half of the resulting vector. This means that after a modification of the input we can reuse the algorithmic pipeline from a non-decimated FFT.

Furthermore, if we compute the DFT of a decimated signal, but want to keep the full frequency resolution of the non-decimated one, then \eqref{eq:FFT_even} and \eqref{eq:FFT_odd} provides an answer to that: one needs to repeat the result from the decimated signal. This \emph{upscaling} operation will be key when designing the network in Section \ref{sec:architecture}.

These operations follow the same principle behind the wide-band butterfly network. If we want to implement \eqref{eq:imaging_cond}, we would need to build a network to process the data at each frequency independently. However, using the argument above one can use the recursive decomposition to process the frequencies jointly. In particular, if we want to process data, say at half frequency, i.e., $\omega/2$, then the complementary low-rank conditions states that $d_x d_y \leq 2\omega^{-1}$. If we suppose, in addition, that the evaluation grid remains constant\footnote{This assumption is a direct consequence of \eqref{eq:imaging_cond}, where  the resolution of the perturbation to be reconstructed is fixed.} then $d_y$ can be twice as large, thus inducing a different factorization. However, as mentioned above, each factor in $H^{\ell}$ factor in the butterfly factorization (see \eqref{eq:butterfly_fact}) down-samples the local representation in $y$, while increasing the resolution in $x$. This means, that after a small modification at the beginning, followed by an upscaling operation similar to the one in \eqref{eq:FFT_even} and \eqref{eq:FFT_odd} when the odd signal is zero, one can reuse the rest of the network, which is idea behind merging the networks to treat the different frequencies jointly at the appropriate scale.

\section{\wbnn Architecture} \label{sec:architecture}

We provide a self-contained overview of the network architecture in this section. This material is tailored towards a machine-learning audience with no prior exposure to the butterfly factorization. Indeed, beyond the salient aspects which we summarize below, implementing \wbnn becomes essentially algorithmic since the network structure and connectivity are determined once the dimensions, i.e., grid size, of the data are specified. \rev{Our discussion and numerical results consider only two-dimensional scattering. In principle the implementation of our architecture in higher dimensions is straightforward as it is essentially prescribed by the corresponding higher-dimensional butterfly factorization. However, we leave the exploration of \wbnn to three-dimensional inverse scattering, and its attendant complications, to future work.}

\begin{figure}[!h]
\centering
\includegraphics[width=\textwidth]{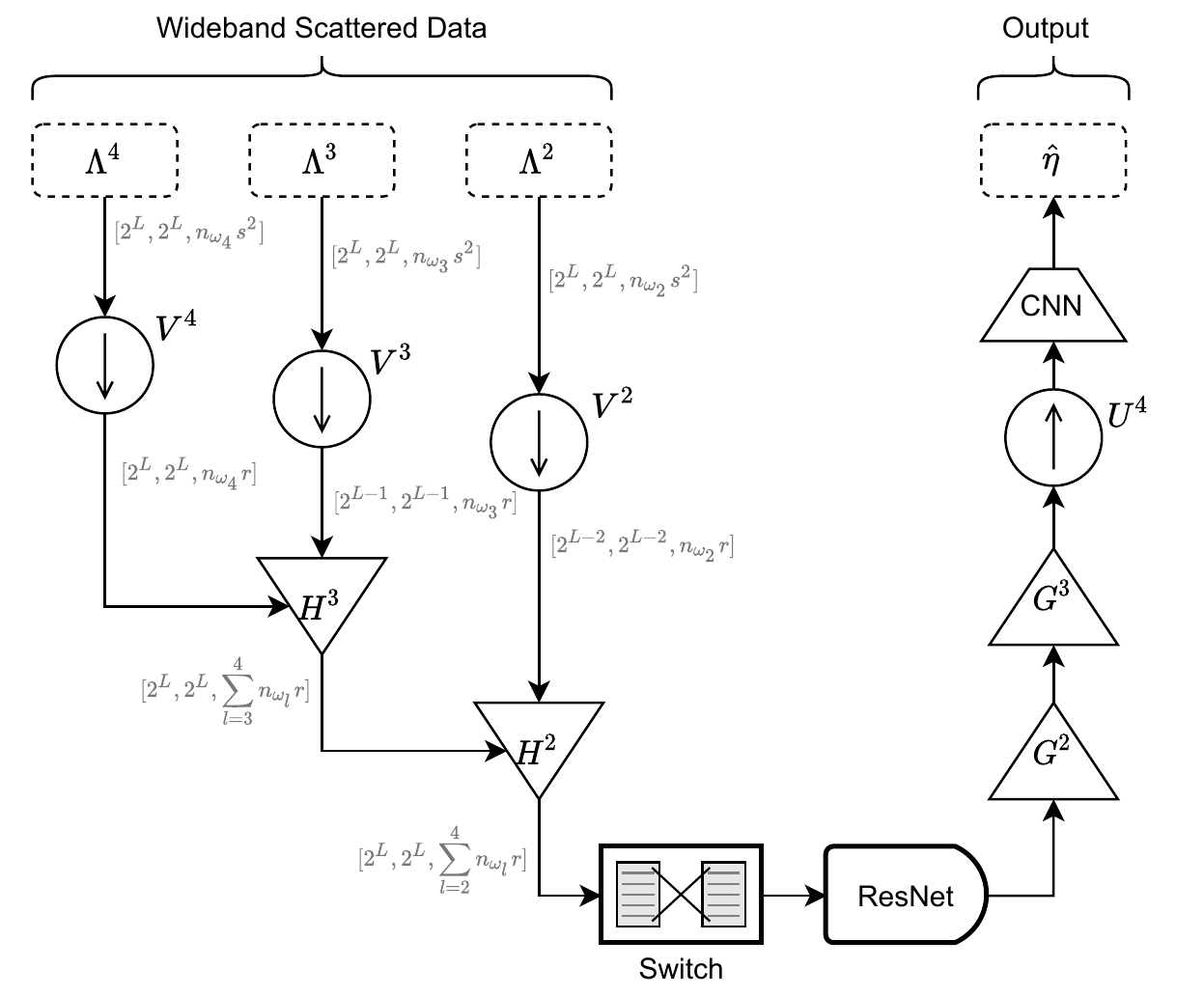}
\caption{Diagram of \wbnn for data with $L=4$ levels, leaf size $s$, and rank $r$. \label{fig:block-diagram}}
\end{figure}

We separate the discussion into the following. In Section \ref{sec:input_data} we define the sampling and formatting of the input data. Section \ref{sec:arch_overview} provides the overarching ideas of the architecture and the layers which comprise it. Details about these layers are further elaborated in their respective sections \S\ref{sec:UV_modules}, \S\ref{sec:HG_modules}, and \S\ref{sec:switch_layer}. Lastly, in Section \ref{sec:network_cost} we discuss the number of parameters (i.e., trainable weights) present in the network. The pseudo-code for \wbnn is provided in Listing~\ref{lst:wbnn} below, whereas the pseudo-codes for the specialized layers $V^\ell$, $H^\ell$, $G^\ell$, and $U^L$ are located in their corresponding subsection\footnote{We use the notation \texttt{LC1D[a,b,c]=LocallyConnected1D(filters=a,kernel\_size=b,strides=c)} throughout.}. Additionally a depiction of the \wbnn architecture for $L=4$ levels is shown in Fig.~\ref{fig:block-diagram}

\begin{lstlisting}[mathescape=true,
floatplacement=tb,label={lst:wbnn},
caption={Pseudo code for the \wbnn, where each module is explained in detail in Sections \ref{sec:HG_modules}, \ref{sec:UV_modules}, and \ref{sec:switch_layer}.} 
]
def wbnn($\Lambda^L$, $\ldots$, $\Lambda^{L/2}$):
  $y$ = $V^L$($\Lambda^L$)
  for $\ell$ in range($L-1$, $L/2-1$, $-1$):
    $y$ = $H^\ell$($y$, $V^\ell$($\Lambda^\ell$))
  $y$ = SwitchResnet($y$)
  for $\ell$ in range($L/2$, $L$, $+1$):
    $y$ = $G^\ell$($y$)
  $y$ = $U^L$($y$)
  $y$ = CNN($y$)
  return y
        
\end{lstlisting}

\subsection{Input formatting} \label{sec:input_data}
We assume the scatterers (discretized over an $n_x \times n_z$ grid) and the scattered data (an $n_\textrm{src} \times n_\textrm{rcv}$ matrix for each frequency~$\omega$) are represented using complete quad-trees with $L$~levels\footnote{We require that $L$ is divisible by 2. This is a minor restriction and can be accommodated by e.g. zero padding of the data or by interpolating the data. While the total depth of both quad-trees must be the same, it is not necessary for them to have the same leaf size. However, for ease of presentation, our discussion focuses exclusively on this case.} with leaf size~$s$. In other words, we require a discretization into $n = 2^Ls$ points for each matrix dimension. The choices of $L$ and $s$ are informed by the inherent wavelengths and sampling frequencies of the inverse problem, and are chosen so that each $s \times s$ voxel of the data matrix~$\Lambda^\omega$ are non-oscillatory, i.e., contain at most a few oscillations.

\begin{figure}[t!]
\centering
\includegraphics[width=0.6\textwidth]{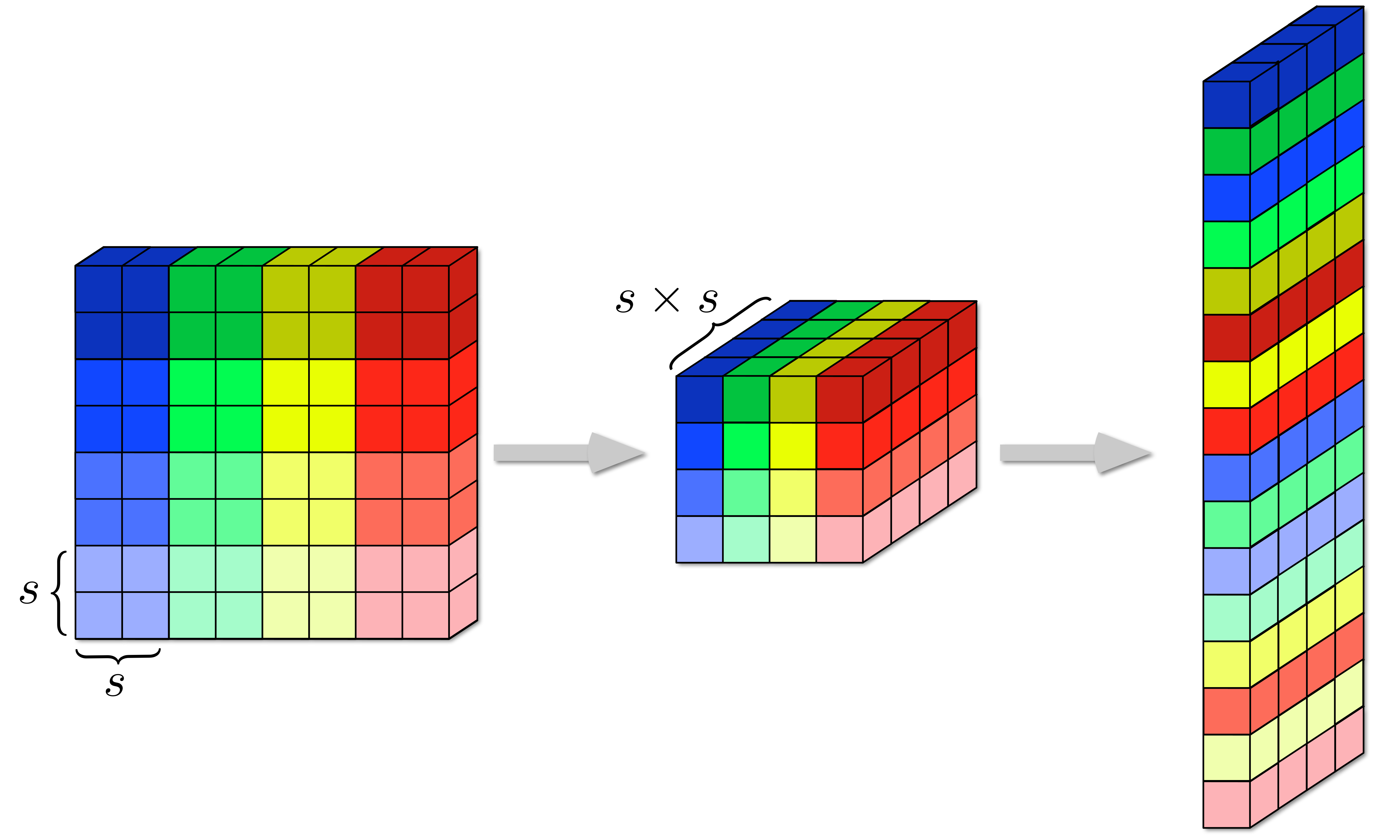}
\caption{ Transformation from the image of dimensions $[2^{\ell}s,~2^{\ell}s]$ to the tensorized form of size $[2^{\ell},~2^{\ell},~s^2]$, and then to the flatted tensor using \rev{Morton} ordering resulting on a tensor of dimensions $[4^L,~s^2]$. \label{fig:input_ordering}}
\end{figure}

Following the Tensorflow convention of \texttt{[height, width, channels]} we reshape these quad-trees into three-tensors of size $[2^L,~2^L,~s^2]$ as shown in Fig.~\ref{fig:input_ordering}. The first two dimensions of the tensor index the geometrical location of the voxels, and the last dimension corresponds to their local vectorial representation. In fact, the data describing the local representation inside each voxel correspond to \emph{channels}. We refer to slices along the height and width dimensions, i.e., the geometrical dimensions, as \emph{patches}. For example, a $1\times1$ patch of data describes slices of the three-tensor with dimension $[1,~1,~s^2]$. As we discuss shortly, at the finest spatial resolution \wbnn operates on $1\times1$ patches, and at the coarsest spatial resolution it operates on $2^{L/2}\times2^{L/2}$ patches. It is convenient to introduce levels $\ell \in [L/2,~L]$ to index the resolution, or equivalently, the size of the contiguous $2^{L-\ell}s \times 2^{L-\ell}s$ sub-matrices in the data matrix that will be processed.

For the purpose of describing our network using linear algebraic operations it is convenient to characterize these three-tensors as equivalently reshaped two-tensors of size $[4^L,~s^2]$. This flattening proceeds according to a natural ordering of quad-trees known as ``\rev{Morton}-ordering'' or ``Z-ordering'', which is depicted in Fig.~\ref{fig:input_ordering}. We refer to \cite{LiYangYing:MultiDimButterflyFact2018} for more details. 


As we discussed in the introduction it is beneficial for the stability of the inverse problem for the input data~$\Lambda^\omega \in \mathbb{C}^{n_\textrm{src} \times n_\textrm{rcv}}$ to be collected from a wideband of frequencies $\omega \in \Omega = [\omega_\textrm{low},~\omega_\textrm{high}]$.
The bandwidth $\omega_\textrm{low}$ and $\omega_\textrm{high}$ is determined from the experimental configuration. For our data assimilation strategy we index this bandwidth with a dyadic partition containing $L/2+1$ intervals: for $L/2 \leq \ell \leq L$ we label the intervals $\Omega^{\ell} = (\omega_\textrm{low} + 2^{L-\ell-1}\Delta\omega,~\omega_\textrm{low} + 2^{L-\ell}\Delta\omega]$ where $\Delta\omega = 2^{-L}(\omega_\textrm{high} - \omega_\textrm{low})$. We assume that within each interval $\Omega^{\ell}$ we probe the medium with $n_\omega^{\ell}$~frequencies, not necessarily equi-spaced, and with slight abuse of notation denote the resulting dataset as $\Lambda^{\ell} \in \mathbb{C}^{n_\textrm{src} \times n_\textrm{rcv} \times n_\omega^{\ell}}$. Following the quad-tree structure we reshape each data tensor $\Lambda^{\ell}$ into a three-tensor of size $[2^{\ell},~2^{\ell},~n_\omega^{\ell} s^2]$ by concatenating all the multi-frequency data collected from bandwidth $\Omega^{\ell}$ along the channel dimension. The input to \wbnn thus consists of the collection of scattering data~$\{\Lambda^{\ell}\}_{L/2 \leq \ell \leq L}$. 

\subsection{Architecture Overview} \label{sec:arch_overview}

\rev{We aim to incorporate the physics of wave propagation into the design of our network by translating analytic properties of the discrete imaging condition~\eqref{eq:imaging_cond} into neural modules. Since the imaging condition is derived by linearization of the partial differential operators in the wave equation, this process should, at minimum, ensure that our network is able to capture the physics of single wave scattering. To that end, for a set of given frequencies $\{\omega_{\ell}\}_{\ell = L/2}^{L}$ we seek an architecture that can emulate the functionality of the imaging algorithm
\begin{equation} \label{eq:wideband_anzats}
\{\Lambda^\ell\} \mapsto \sum_{\ell=L/2}^{L} \alpha(\omega_{\ell}) \left ( \left(F^{\omega_{\ell}} \right )^* F^{\omega_{\ell}} + \epsilon I  \right )^{-1} \left(F^{\omega_{\ell}} \right )^*  \Lambda^{\ell}.
\end{equation}
We emphasize, however, that we are ultimately interested in applications of \wbnn to data beyond the Born single scattering regime associated with the imaging condition.}

We leverage the following analytic properties of the imaging condition. First, as originally elucidated in \cite{Khoo_YingSwitchNet:2019}, we recognize that for each frequency~$\omega$ the regularized normal operator~$\left ( \left(F^{\omega} \right )^* F^{\omega} + \epsilon I \right )^{-1}$ corresponds to a translation invariant operator. Second, we also recognize that the operator $F^\omega$ describes a generalized Fourier operator which is amenable to a butterfly factorization. In other words, after suitable discretization the operator $F^\omega$ admits a matrix decomposition viz.,
\begin{equation}
F^\omega  = U^{L} G^{L-1} \ldots G^{L/2} S^{L/2} H^{L/2} \ldots H^{L-1} V^{L}.
\label{eq:butterfly_net}
\end{equation}
In the traditional linear setting of the discrete imaging condition with \rev{Morton}-flattened data we have $U^L \in \mathbb{C}^{4^Ls^2 \times 4^Lr}$,~$V^L \in \mathbb{C}^{4^Lr \times 4^Ls^2}$, and all other remaining matrix factors of dimension $4^Lr \times 4^Lr$. Most importantly, each matrix factor in the butterfly decomposition has a sparsity pattern that is informed by analytic considerations of the wave kernel. These sparsity patterns in the matrix become equivalent to a block diagonal operator after specific permutation of either the columns or the rows.

\wbnn utilizes these two insights to replace the functionality of $\left ( \left(F^{\omega} \right )^* F^{\omega} + \epsilon I \right )^{-1}$ and $F^\omega$ by analogous neural modules. We translate the butterfly decomposition for the generalized Fourier operator $F^\omega$ by replacing each matrix factor (e.g. $U^L$, $V^L$,~\ldots) by neural network layers\footnote{For ease of comparison we retain the transpose $\cdot^*$ in the naming convention of our network but note that transposition is no longer meaningfully defined in our new non-linear setting.}. In this setting the permutation and sparsity structure of the butterfly matrix factors inform the inter-layer network connectivity (i.e., the network topology), and the matrix entries themselves become trainable weights in the network. The information processed by these layers are ultimately sent data into a \textsc{CNN} module, which are well adapted to capture translation invariant operators, and thus  mimics the effects of the regularized pseudo-inverse in sharpening the estimate coming from the imaging condition.



If only monochromatic data are considered, e.g. using solely scattering data $\Lambda^{L} \in \mathbb{C}^{n\times n \times n_\omega^{L}}$ obtained by probing the medium at only $n_\omega^{L}=1$ frequency, then the network just described is equivalent to other butterfly-based networks \textsc{BNet} \cite{Yingzhou2018} or \textsc{SwitchNet} \cite{Khoo_YingSwitchNet:2019}. However, rather than replacing each $\omega$-dependent operator in the imaging condition \eqref{eq:wideband_anzats} with individual butterfly-based networks, \wbnn instead aims to more efficiently assimilates multi-frequency data with the following modifications:
\begin{enumerate}[label=(\roman*)]
\item We exploit the connection between spatial resolution and frequency in wave-scattering problems by processing data only at their relevant length scales. We note that each $H^\ell$ layer, analogous to their butterfly factorization namesakes, processes data over voxel patches of size $2^{L-\ell} \times 2^{L-\ell}$, i.e., the effective length scales at this layer are of order $2^{L-\ell}$. As a result, the dispersion relation in wave-scattering suggests that data from bandwidth $\Omega^{\ell}$ are most informative at this length-scale\footnote{The $\{G^{\ell}\}$ layers have a similar multi-resolution property. This suggests that data from bandwidth $\Omega^{\ell}$ should also be fed into $G^{\ell}$ similar to the U-Net \cite{U-Net} architecture; however, numerical results demonstrate that this additional complexity is unnecessary.}, and thus we feed in data accordingly. This strategy of dyadically partitioning the bandwidth to localize spatial information is also employed by the Cooley-Tukey FFT algorithm to achieve quasi-linear time complexity \cite{Cooley_Tukey:1965}; in our setting this strategy affords us significant reductions in the number of trainable weights in the network. 

\item In addition to the switch permutation layer, we also introduce non-linearities into the network using residual network which we call the \textsc{switch-resnet} layer. Information from the entire bandwidth of data is thus processed at this layer. These non-linearities, in theory, extend the functionality of \wbnn beyond the limitation the discretized imaging condition; namely, the implicit assumption of Born single scattering. Furthermore, non-linear combinations of wideband data are known to be a strict requirement of super-resolution imaging~\cite{Donoho:super_resolution1992}, and therefore potentially enabling \wbnn image estimates to achieve resolutions below the Nyquist limit.
\end{enumerate}

\noindent In the following sections we elaborate on the specific details of each specialized butterfly-network layer in Alg.~\ref{lst:wbnn}. 

\subsection{\texorpdfstring{$U^{L}$ and $V^{\ell}$ layers}{UL and Vell layers}} \label{sec:UV_modules}
\begin{figure}[t!]
\centering
\includegraphics[width=0.25\textwidth]{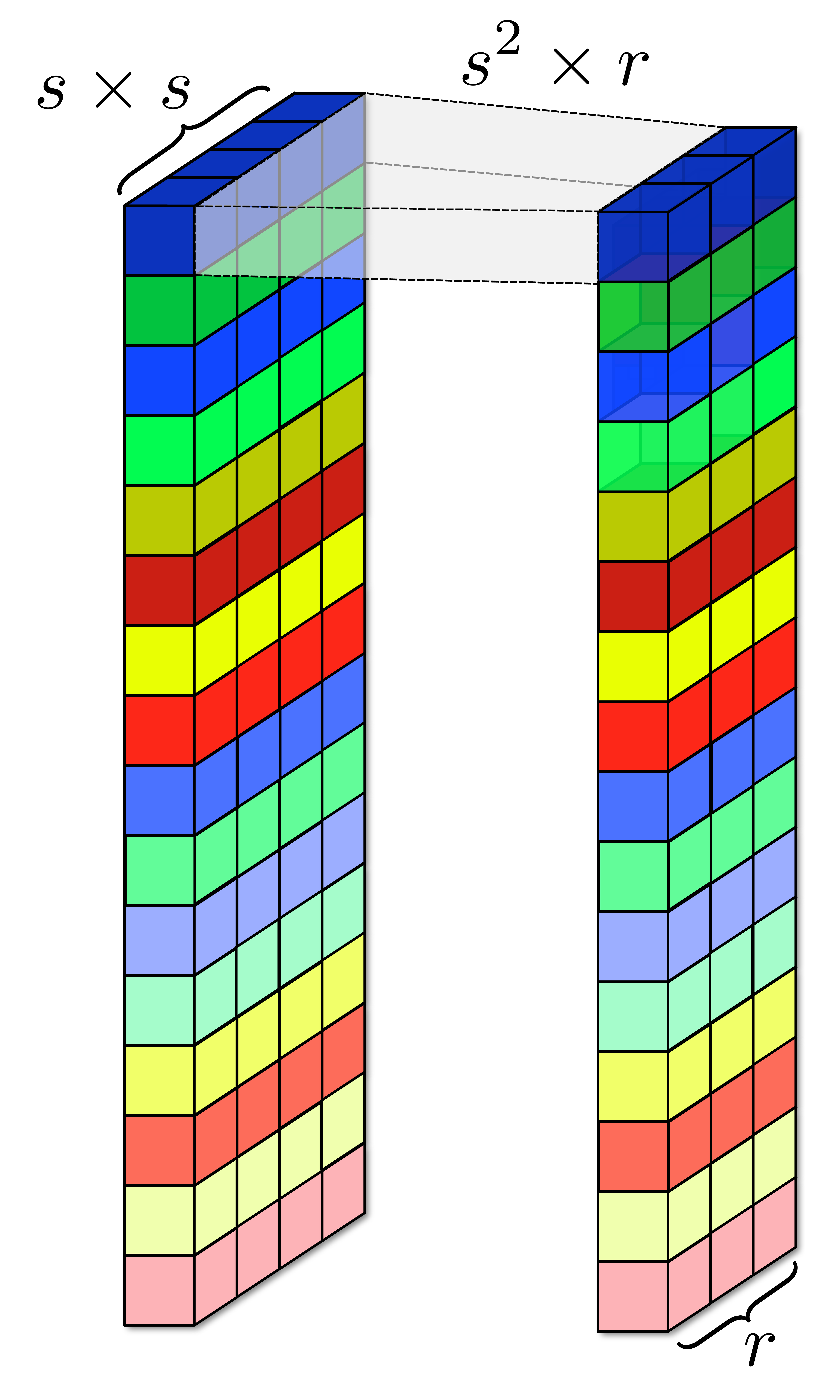}
\caption{Sketch of the compression carried in the $V^{L}$ layer, from the points contained in a leaf of size $s\times s$ (see Fig.~\ref{fig:input_ordering}) to a local representation of rank $r$. The grey polygon represent the connections between the two layers. \label{fig:V_L}}
\end{figure}

In the traditional numerical analysis setting the butterfly matrix factor $V^{\ell}$ in \eqref{eq:butterfly_net} represents a block diagonal matrix with block size $r \times s^2$. This operator takes input data (viewed as a complete quad-tree) and compresses leaf nodes at level~$L$, each with $s \times s$ degrees of freedom, into $1\times1$ patches with $\sqrt{r} \times \sqrt{r}$ degrees of freedom; this process is depicted in Fig.~\ref{fig:V_L}. Similarly, the $U^{L}$ matrix factor in \eqref{eq:butterfly_net} is also block diagonal but instead with block sizes of $s^2 \times r$. This operator thus ``samples'' the local representation of dimension~$r$ back to its nominal dimensions of $s \times x$. In both instances the compression/decompression is essentially lossless provided the number of levels~$L$ is properly adapted to the probe frequency~$\omega$. We emphasize again that this follows as a consequence of the \emph{dispersion relation} in wave-scattering: provided these parameters are chosen correctly, then over $s \times s$ length scales the data are non-oscillatory (i.e. sub-wavelength) and therefore admits a low-rank representation with rank~$r$. 

\wbnn also exploits this relation between spatial resolution and frequency. However, a key point of departure from the butterfly factorization is that here the input data are wideband and thus contains multiple length scales (wavelengths). This motivates the introduction of auxiliary layers $V^{\ell}$ for $L/2 \leq \ell \leq L$ whose inputs are assumed to be sampled from bandwidth $\Omega^{\ell}$. Each $V^{\ell}$~layer compresses the input data \emph{at level}~$\ell$ such that nodes with $2^{L-\ell}s \times 2^{L-\ell}s$ degrees of freedom are mapped into $1\times1$ patches with $\sqrt{r} \times \sqrt{r}$ degrees of freedom; this also has the interpretation of spatial downsampling. Note that the dyadic scaling in the definition of $\Omega^{\ell}$ is critical in maintaining the balance between spatial resolution and frequency. 

When the input data $\Lambda^{\ell}$ from bandwidth $\Omega^{\ell}$ are represented as a three-tensor of dimension $[2^{\ell},~2^{\ell},~n_\omega^{\ell}]$, each $V^{\ell}$ layer can be implemented as a \texttt{LocallyConnected2D} layer in Tensorflow with $rn_\omega^{\ell}$~channels and both the kernel size and stride as $2^{L-\ell}\times 2^{L-\ell}$.  The $U^{L}$ layer can also be implemented as \texttt{LocallyConnected2D} layer with rank~$s^2$ and $1\times1$ kernel size and stride; the input to this layer is assumed to be of dimension $[2^{\ell},~2^{\ell},~c]$ with $c$~input channels. For completeness, we provide the implementation of these layers in Algs~\ref{lst:Vl} and \ref{lst:U} for input data that is Morton-flattened. Furthermore, note the pseudo-code also details the processing when input data contain both real and imaginary components.

\begin{lstlisting}[mathescape=true,
caption={Pseudo code for the $V^{\ell}$ module for Morton-flattened data.},label={lst:Vl}
]
def $V^\ell$($X$):
  # input:  [?, 2, $4^\ell$, $4^{L-\ell}s^2n_{\omega_\ell}$] 
  # output: [?, 2, $4^L$, $rn_{\omega_\ell}$] 
  $y_\textrm{re}$, $y_\textrm{im}$ = $X$[:,0,:,:], $X$[:,1,:,:] 

  $x_\textrm{re}$ = LC1D[$rn_{\omega_\ell}$,1,1]($y_\textrm{re}$) + LC1D[$rn_{\omega_\ell}$,1,1]($y_\textrm{im}$)
  $x_\textrm{im}$ = LC1D[$rn_{\omega_\ell}$,1,1]($y_\textrm{re}$) + LC1D[$rn_{\omega_\ell}$,1,1]($y_\textrm{im}$)

  return tf.stack([$x_\textrm{re}$, $x_\textrm{im}$], axis=1)


\end{lstlisting}
\begin{lstlisting}[mathescape=true,
caption={Pseudo code for the $U^L$ module for Morton-flattened data.},label={lst:U}
]
def $U^L$($X$):
  # input:  [?, 2, $4^L$, $\cdot$]
  # output: [?, 2, $4^L$, $s^2$]
  $y_\textrm{re}$, $y_\textrm{im}$ = $X$[:,0,:,:], $X$[:,1,:,:]

  $x_\textrm{re}$ = LC1D[$s^2$,1,1]($y_\textrm{re}$) + LC1D[$s^2$,1,1]($y_\textrm{im}$)
  $x_\textrm{im}$ = LC1D[$s^2$,1,1]($y_\textrm{re}$) + LC1D[$s^2$,1,1]($y_\textrm{im}$)

  return tf.stack([$x_\textrm{re}$, $x_\textrm{im}$], axis=1)
    
\end{lstlisting}

\textbf{Remark:} A major application of the butterfly factorization is for applying FIOs in linear-time complexity; the rank~$r$ then depends on the error tolerance but generally requires that $r \ll s^2$. This represents a significant philosophical difference in how in $r$ is determined in our machine learning setting -- it does not matter if $r \geq s^2$ so as long as the learned model achieves its intended task. Nevertheless, our numerical results in \S\ref{sec:training_curves} demonstrate that \textit{(i)} it suffices to choose $r \ll s^2$ and, moreover, that \textit{(ii)}~generalization is largely insensitive to the choice of $r$.

\subsection{\texorpdfstring{$H^{\ell}$ and $G^{\ell}$ layers}{Hell or Gell layers}} \label{sec:HG_modules}
\begin{figure}[t!]
\centering
\includegraphics[width=0.35\textwidth]{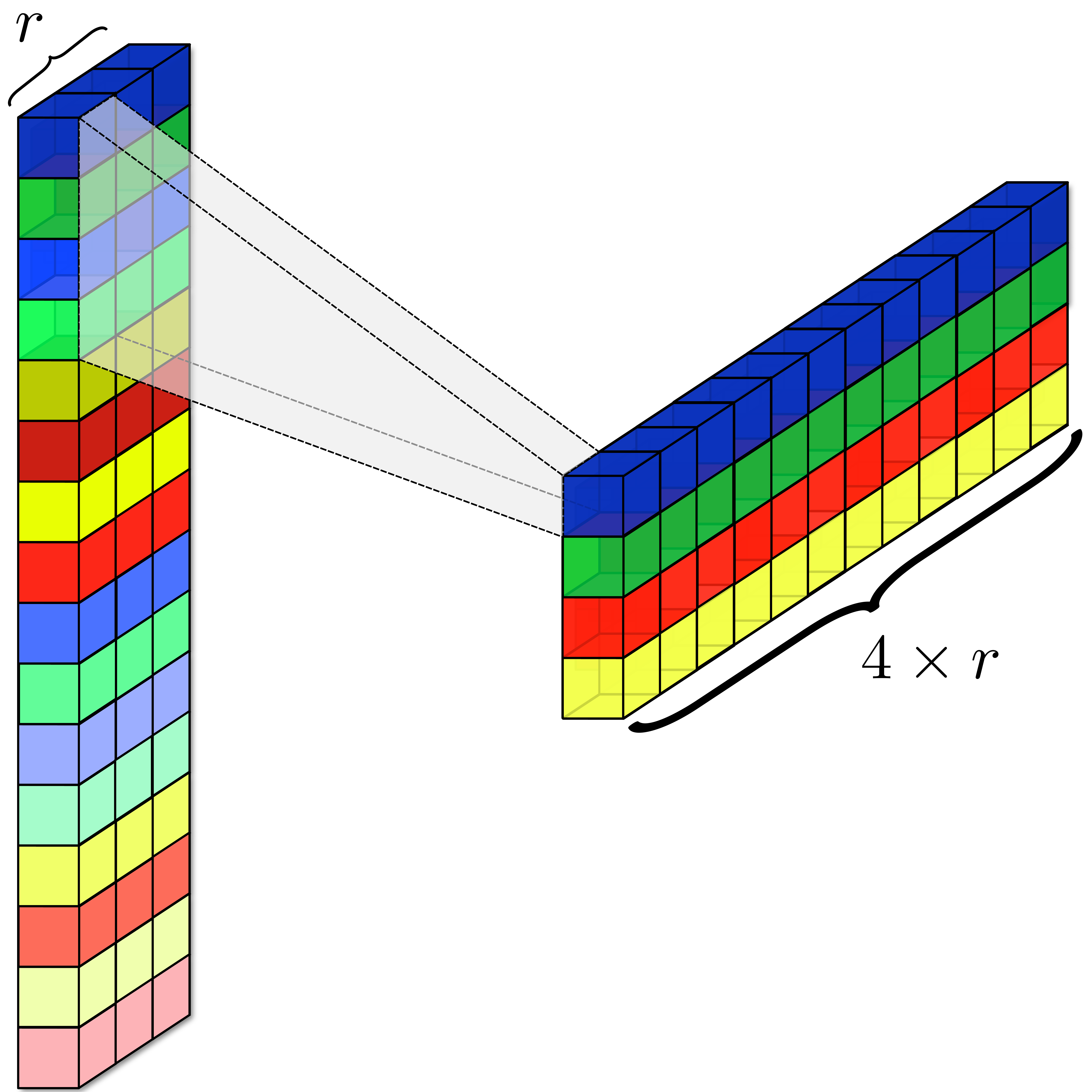}
\caption{Sketch of the application of the $H^{\ell}$ layer. The layer decimates by a factor of four the number of neurons in the spatial dimension, while increasing four times the number of channels. From Fig.~\ref{fig:input_ordering} we can observe that the decimation is equivalent to decimate by a factor two in each of the first two dimensions, which follows from the Z-ordering. \label{fig:H_ell}}
\end{figure}

The $H^{\ell}$ and $G^{\ell}$ factors in \eqref{eq:butterfly_net} continue the theme of multi-scale processing. When viewed as matrices, both $H^{\ell}$ and $G^{\ell}$ are block diagonal with block size $4^{L-\ell}r \times 4^{L-\ell}r$. Equivalently, when the input is formatted as a complete quad-tree, this implies both are \textit{local operators} which process the nodes on the tree at length scale $l$ to map each $2^{L-\ell}s \times 2^{L-\ell}s$ patches. Within each block there is further structure to the operators, as Figure \ref{fig:H_ell} demonstrates. For each $H^{\ell}$ each sub-block has the interpretation of \emph{aggregating} information, whereas each $G^{\ell}$ achieves the dual task of \emph{spreading} information. We stress, however, that the action of this is entirely local within each patch. In either case, the key observation is that by permuting each node following a set pattern each operator becomes block-diagonal with block size $4^\ell$, for all $L/2 \leq \ell \leq L$. The specific permutation pattern $\pi_\ell$ enabling this matrix partitioning is discussed in Appendix~\ref{sec:perm_switch_idx}.

In our \wbnn adaptation, each $G^{\ell}$ layer directly mimics the behaviour of their counterparts and can be implemented using the \textsc{LocallyConnected2D} layer with $4 \times 4$ kernel sizes and stride $4$. The number of channels is chosen to be $\sum_{i=\ell+1}^{L} rn_{\omega_i}$ for symmetry. 

However, note that our $H^{\ell}$ layers require modification on account of our data assimilation strategy to inject information at their correct length scales. As such, these layers process two inputs: one the output of the $V^{\ell}$ layer of dimension $[2^{L-\ell}, 2^{L-\ell}, rn_{\omega_{\ell}}]$, the other the output from the previous layer of dimension $[2^{\ell}, 2^{\ell}, c]$ for some channel size $c$. To process the dimensions of both we first upscale each patch with redundant information to convert the data into $[2^{\ell}, 2^{\ell}, rn_{\omega_{\ell}}]$. This is then concatenated with the other input to form a tensor of size $[2^{\ell}, 2^{\ell}, c+rn_{\omega_{\ell}}]$. Note that the ordering of the concatenation along the channel dimension does not matter so as long as it is performed consistently.

Alg~\ref{lst:H} provides a pseudo-code implementation of the $H^\ell$ layer when using Morton-flattened inputs. 

\begin{lstlisting}[mathescape=true,
floatplacement=tb,label={lst:H},
caption={Pseudo code for the $H^\ell$ module for Morton-flattened data.}
]
def $H^\ell$($X$, $W$):
  # input $X$:  [?, 2, $4^\ell$, $\cdot$] from $V^\ell(\Lambda^\ell)$
  # input $W$:  [?, 2, $4^L$, $\cdot$] from previous layer
  # output: [?, 2, $4^L$, $r \sum_{i=\ell}^L n_{\omega_i}$]  
   
  $\widetilde{X}$ = UpSampling2D($X$)

  $X$ = tf.stack([$X$, $\widetilde{X}$], axis=-1)

  $y_\textrm{re}$, $y_\textrm{im}$ = $X$[:,0,:,:], $X$[:,1,:,:]

  # permute patches according to $\pi_\ell$ before
  # applying block diagonal transformations
  $y_\textrm{re}$ = $y_\textrm{re}$[:,$\pi_\ell$,:]
  $y_\textrm{im}$ = $y_\textrm{im}$[:,$\pi_\ell$,:]

  $x_\textrm{re}$ = LC1D[4$r \sum_{i=\ell}^L n_{\omega_i}$,4,4]($y_\textrm{re}$) + LC1D[4$r \sum_{i=\ell}^L n_{\omega_i}$,4,4]($y_\textrm{im}$)
  $x_\textrm{im}$ = LC1D[4$r \sum_{i=\ell}^L n_{\omega_i}$,4,4]($y_\textrm{re}$) + LC1D[4$r \sum_{i=\ell}^L n_{\omega_i}$,4,4]($y_\textrm{im}$)

  return tf.stack([$x_\textrm{re}$, $x_\textrm{im}$], axis=1)
\end{lstlisting}

\subsection{\textsc{Switch-Resnet} layer} \label{sec:switch_layer}
\begin{figure}[t!]
\centering
\includegraphics[width=0.6\textwidth]{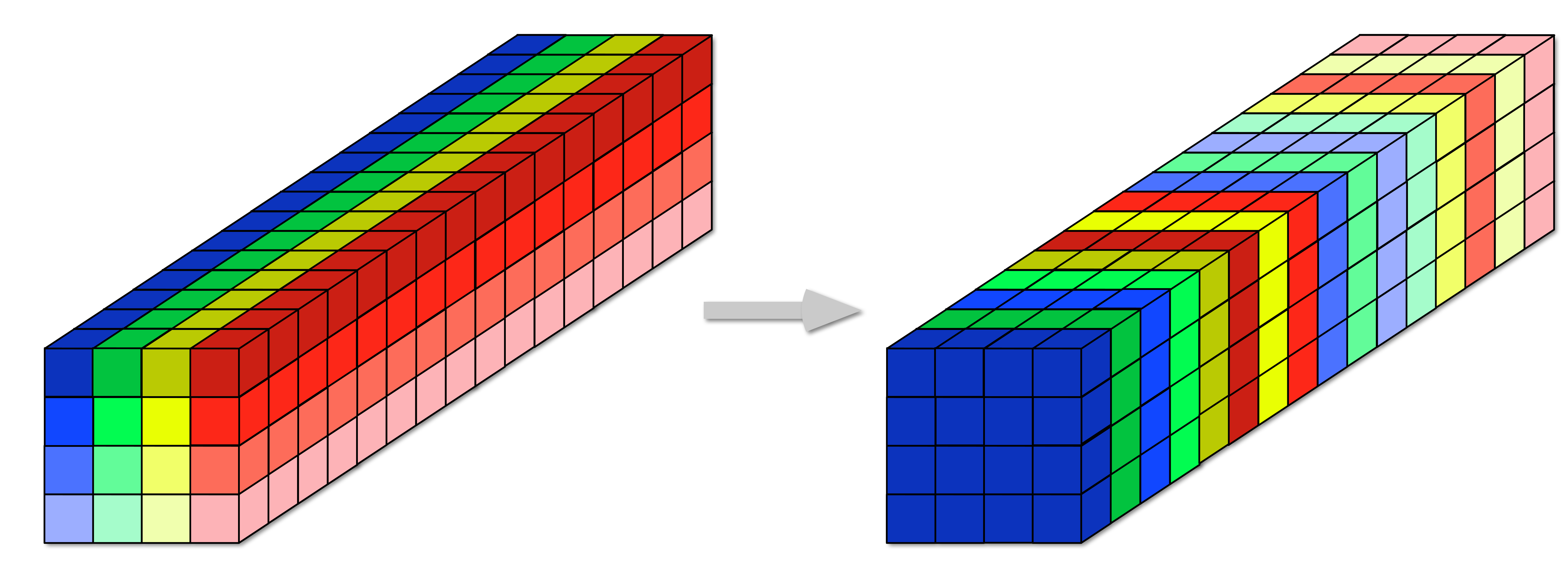}
\caption{\rev{Visualization of the transformation by the switch permutation layer for an example with $r=1$ and $L=4$. For this configuration the input and output are both tensors of dimension $[2^{L/2}, 2^{L/2}, r 4^{L/2}] = [4,4,16]$. The local information contained at each geometric position (equivalently, the channel information) is distributed globally according to the switch permutation pattern~$\pi_\textrm{switch}$. \label{fig:switch_ordering}}}
\end{figure}

We retain the permutation pattern of the switch layer as this is responsible for capturing the inherent non-locality of wave scattering (e.g. a point scatterer generates a diffraction pattern that is measured by all receivers in our geometry). We illustrate this pattern in Fig.~\ref{fig:switch_ordering}, and the specific description of the permutation indexing~$\pi_\textrm{switch}$ can be found in Appendix~\ref{sec:perm_switch_idx}.

The input to this level serves as a condensed representation of the measured data. It is at this level that we non-linearly process the multi-frequency dataset; we speculate that this also essential in facilitating the model to produce super-resolved images. We achieve this using a residual network to refine each channel locally following each resnet unit. The pseudocode is provided in Alg.~\ref{lst:switch}.

\begin{lstlisting}[mathescape=true,
floatplacement=tb,label={lst:switch},
caption={Pseudo code for the \textsc{Switch-Resnet} module for Morton-flattened data.}
]
def SwitchResnet(X):
  # apply switch local-to-global permutation
  $y$ = $X$[:,:,$\pi_\textrm{switch}$,:]

  # non-linear synthesis
  # apply switch permutation
  for k in range($N_\textrm{resnet}$):
    $y$ = LC1D[r,1,1](ReLu(LC1D[r,1,1]($y$)) + $y$
    if k < $N_\textrm{resnet}$:
      $y$ = ReLu($y$)

\end{lstlisting}

\subsection{\wbnn Parameter Count} \label{sec:network_cost}
An estimate of how the number of parameters (i.e. trainable weights or \rev{degrees of freedom (d.o.f.))} scales is

\[
\textrm{d.o.f.}(\wbnn) \approx 4^{\ell}r^2 \left( \sum_{l=L/2}^{\ell} n_{\omega_{\ell}}^2 + \sum_{l=L/2}^{\ell} \left(\sum_{i=l}^{\ell} n_{\omega_i}\right)^2 \right).
\]
When only a single frequency is sampled in each sub-band, i.e. $n_{\omega_{\ell}}=1$ for all $l$, then this total becomes $\mathcal{O} \left ( N(\log N + \log^3 N)\right)$. Note this is essentially linear in the total degrees of freedom in the data ($N$) up to poly-logarithmic factors. Furthermore, note if na\"ively $L$ separate single channel \wbnn networks were used to compute \eqref{eq:imaging_cond} this would correspond to complexity $ \mathcal{O} \left ( N\log N^2 \right)$; the multi-frequency assimilation only exceeds this with mild oversampling by a logarithmic factor.

Lastly, we note the effect of the partitioning of the frequencies.  If all the frequencies were ingested at length scale $L$ then the scaling becomes $\mathcal{O}\left (N(\log N^2 + \log N^3) \right)$. While to leading order this presents the same asymptotic scaling, in terms of practical considerations this presents as substantial increase in the number of trainable parameters.

\section{Numerical Results}
\label{sec:numerical_results}

Synthetic data were generated using numerical finite differencing for \eqref{eq:scattering} over the computational domain $[-0.5, 0.5]^{\otimes 2}$. The domain was discretized with an equispaced mesh of $n_x=80$ by $n_z=80$ points which corresponds to a quad-tree partitioning into $L=4$ levels with leaf size $s=5$. Training data were generated using a second-order finite difference scheme while \textit{testing} data were computed with fourth-order finite differences. The use of higher quality simulations for testing serves to validate that \wbnn predictions do not depend on computational artifacts such as e.g. numerical dispersion. The radiating boundary conditions for Eq~\ref{eq:scattering} were implemented using perfectly matched layers (PML) with a quadratic profile with intensity 80 \cite{Berenger:PML}. The width of the PML was chosen to span at least one wavelength at the lowest frequency. 

Unless specified otherwise the dataset consisted of $n_f=3$ source frequencies at $2.5$, $5$, and $10$ Hz. In a homogeneous background with velocity $c_0 = 1$ this corresponds to $8$ points-per-wavelength (PPW) at the highest frequency. Receivers were located at equi-angular intervals around a circle of radius $r=0.5$ with the recorded data computed by linearly interpolating the scattered field. We used $N_{\textrm{rcv}} = 80$ receivers and sources for all experiments. For a homogeneous background the direct wave is given analytically (see \eqref{eq:scattering}). In these instances the directions of arrival $\mathbf{s} \in \mathbb{S}^1$ were aligned with the receiver geometry, i.e. incident from $80$~equiangular directions. However, for inhomogeneous media the direct waves had to be computed numerically. This was achieved by using numerical Dirac deltas as source functions. These sources were localized on a circle of radius $r=1$ at $80$ equiangular intervals and the computational domain was extended to $[-1,1]^{\otimes 2}$ using the same grid spacing $\Delta x$ and $\Delta z$ as before. The resulting scattered field was computed by differencing the solutions to \eqref{eq:scattering} with and without scatters. The acquisition geometry was fixed for all frequencies. 

Scatterers were selected from a dictionary of simple, convex, geometric objects such as squares, triangles, and \rev{Gaussian} bumps. The characteristic lengths of the square and triangular scatterers were measured with respect to their base, rather than the diameter of the smallest enclosing ball, whereas the characteristic length of the \rev{Gaussian} was taken to be its standard deviation. In each data point the number of scatterers was determined by uniformly sampling from $\{2,3,4\}$ objects, and their locations were uniformly distributed inside a circle of radius~$r=0.35$. No restrictions were enforced against overlapping scatterers. In all experiments the amplitude of each scatterer was fixed to~$0.2$; we leave to future work how the training data can be augmented to account for variations in amplitudes.

\wbnn was implemented in Tensorflow \cite{tensorflow2015} and trained with the pixel-wise sample loss function
\begin{equation}
\label{eq:loss}
\sum_{x=\textrm{pixel in image}} \left \| (K_\textrm{high} * \eta)(x) - \wbnn[\Lambda^{L}, \ldots, \Lambda^{L/2}](x) \right \|_2^2,
\end{equation}
where $\eta$ denotes the sample realization of the scatterer wavefield and $\{\Lambda^{\ell}_{s,r}\}_{L/2\leq l \leq L}$ the partitioned multi-frequency data. This objective function was chosen to promote the recovery of an image that is \textit{smoother} than the true numerical solution by a factor of a two-dimensional convolution with high-pass filter $K_\textrm{high}$. Critically we still remain in the super-resolution regime when the support of filter $K_\textrm{high}$ is significant smaller than the Nyquist limit of $\lambda_{min}/2$\footnote{The ratio of these two quantities is the so-called \textit{super-resolution factor}.} as the smoothed image still contains sub-wavelength features. This strategy was inspired by the work of \cite{Cands2013} who relied on this insight for theoretical proofs on recoverability limits in super-resolution. In our experiments we selected $K_\textrm{high}$ to be a \rev{Gaussian} kernel with characteristic width of $0.75$~grid points (compare this the diffraction limit in our bandwidth of $4$~pixels). This smoothing was observed to be integral in promoting stable training dynamics. We also report the image-wise relative error
\begin{equation}
\label{eq:relloss}
\frac{ \displaystyle \left \| K_\textrm{high} * \eta - \wbnn[\Lambda^{L}, \ldots, \Lambda^{L/2}]\right \|_2^2}{\displaystyle \left \| K_\textrm{high} * \eta \right \|_2^2}.
\end{equation}
Note that we do not normalize the norms in either \eqref{eq:loss} or \eqref{eq:relloss} by the grid lengths $\Delta x$ and $\Delta z$.

The dataset was split into $21000$ training points  and $4000$ testing points\footnote{a single ``data point'' has dimension $n_x \times n_z \times n_f$}, respectively, with batch size~$32$. Note, in comparison, an instance of \wbnn with $N_{\textrm{CNN}}=3$ convolutional layers and $N_{\textrm{RNN}}=3$ residual layers contains $~200000$ trainable parameters meaning our models are still in the massively over-parameterized regime.  Unless specified otherwise the testing set follows the same distribution (e.g. scatterer types) as the training set. The initial learning rate (i.e. step size) was universally set to 5e-3 across all experiments. The learning schedule was set according to Tensorflow's \cite{tensorflow2015} implementation of \texttt{ExponentialDecay} with a decay rate of $0.95$ after every $2000$~plateaus steps with stair-casing. We chose the Adam optimizer \cite{kingma2015adam} and terminated training after $150$~epochs. No special initialization strategy was required and the network weights were randomly initialized with \texttt{glorot\_uniform} -- we did not observe the training instabilities with random initialization that were thoroughly documented in \cite{Butterfly-Net2} for general butterfly networks. All computations were done with \texttt{float32} half-precision. Note that no effort was taken to optimize these hyper-parameters using an external validation set.

\subsection{Homogeneous Background}
\label{sec:homogeneous_bkgd}

\begin{figure}[htp!]
     \centering
     \includegraphics[width=\textwidth,trim={00mm 2.8cm 00mm 00mm},clip]{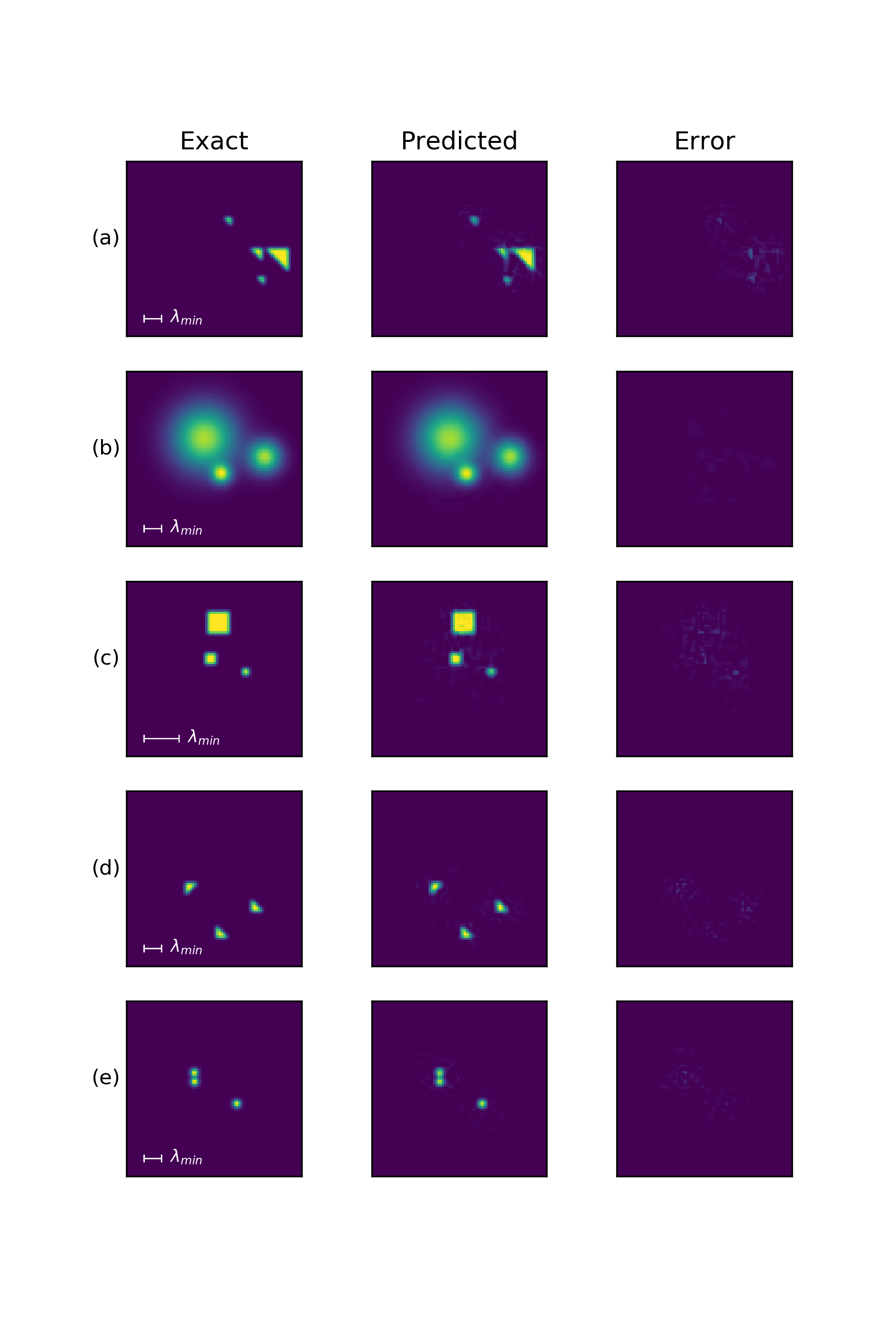}
     \caption{Visualization of \wbnn predictions on a testing set. The first column is the exact solution, the second column the output of \wbnn, and the third column the point-wise error. The colour scales in each row are normalized with respect to the first column. (a) with $\triangle$~(3,5,10). (b) same as above but with the \rev{Gaussian} dataset. (c) heterogeneous squares but with a lower bandwidth (1.25, 2, and 5 Hz) so the effective wavelength is $16$~PPW. (d) rotated dictionary. (e) \rev{Gaussian} scatterers with characteristic length~$1$.}
     \label{fig:homogeneous}
 \end{figure}

In this section we present numerical results for \wbnn models trained with scattered data that propagated through a known homogeneous background medium of wavespeed $c_0 = 1$. Each row of Figure~\ref{fig:homogeneous} depicts \wbnn predictions on testing data across a variety of scatterer configurations. Except for Figure~\ref{fig:homogeneous}c the data were sampled from the bandwidth of 2.5, 5 and 10~Hz which implies a limiting wavelength of $8$~points per wavelength (PPW). Figures~\ref{fig:homogeneous}a and \ref{fig:homogeneous}b involve a multi-scale dictionary of scatterers with characteristic lengths ranging from $3$, $5$, and $10$ pixels; these correspond to the sub-wavelength, wavelength, and super-wavelength regimes, respectively. We observe that \wbnn correctly localizes each scatterer in addition to resolving sub-wavelength features such as e.g. the corners of the triangles. Figure~\ref{fig:homogeneous}d similarly depicts a heterogeneous dictionary but with rotated triangles of fixed side-length~$5$ pixels. In Figure~\ref{fig:homogeneous}c the same experiment was repeated but with a bandwidth that was shifted to $1.25$, $2.5$ and $5$ Hz so that the limiting wavelength increases to $16$~PPW; in this regime all scatterers are sub-wavelength. Nevertheless, \wbnn still produces images that are qualitatively comparable to the higher bandwidth experiments. This suggests that our algorithm has a high super resolution factor. For completeness,
we include results in Figure~\ref{fig:homogeneous}e for point scatterers that were originally proposed for super-resolution by Donoho \cite{Donoho:super_resolution1992}.

Table~\ref{tab:numerical_results} summarizes the training and testing loss for various scatterer configurations. Each row corresponds to a separate experiment with triangular ($\triangle$), square ($\square$), or \rev{Gaussian} ($\bigcirc$) scatterers. The numbers in the parentheses correspond to the characteristic length, in pixels, with multiple numbers indicating a multi-scale dataset. 

Several trends can be observed from this table. In all configurations there is no evidence of over-fitting; indeed, the generalization gap, defined to be the difference between the testing and training errors, is on average less than an order of magnitude. Furthermore, both qualitatively and quantitatively there is no significant difference between datasets with a fixed characteristic length versus the multi-scale datasets. This demonstrates robustness to the choice of the scatterer dictionary. However, we observe that \rev{Gaussian} scatterers outperform other shapes across all metrics, perhaps owing to their smoothness. Overall, the pixel-wise error in testing tends to decrease with decreasing length scale; we conjecture the exact scaling may depend on the perimeter to area ratio of the polygons.

\subsubsection{Effect of Switch Layer}
\begin{figure}[htp!]
     \centering
     \includegraphics[width=\textwidth,trim={00mm 3cm 00mm 00mm},clip]{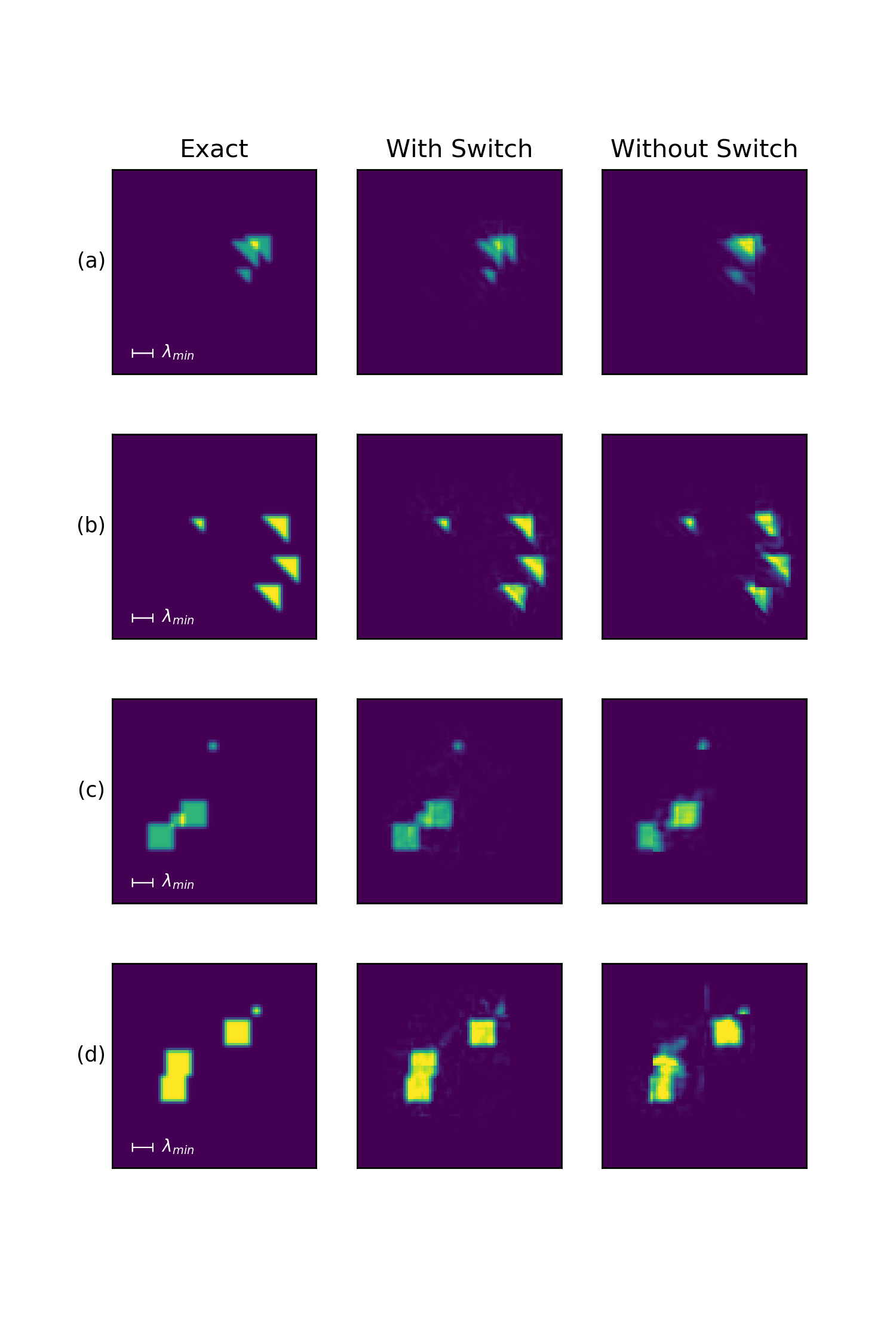}
     \caption{Visualization the effect of removing the switch permutation layer. The colour scale of each row is normalized to the first column. We observe that while \wbnn-without-switch manages to localize the scatterers, it is unable to fully resolve all sub-wavelength features.}
     \label{fig:effect_of_switch}
 \end{figure}

In Section~\ref{sec:architecture} we emphasized the importance of the \textsc{switch} permutation pattern in representing the local-to-global physics of wave scattering. Figure~\ref{fig:effect_of_switch} corroborates this claim by comparing the predictive ability of \wbnn models trained with and without the inclusion of the \textsc{switch} permutation layer. Both models contain the same number of trainable weights, and all other configurations were held equal. 

Figure~\ref{fig:effect_of_switch} demonstrates that the predictions \emph{without} the permutation layer are of noticeably poorer quality. However, the switch-less configuration manages to localize scatterers and even reproduces sub-wavelength features to an extent, particularly when the scatterers are well separated as in Figure~\ref{fig:effect_of_switch}(b). However, Figures~\ref{fig:effect_of_switch}(c) and (d) exposes the deficiencies of this model in the presence of overlapping scatterers, i.e., in the super-resolution regime where scatterers are separated by sub-wavelength distances. We observe that these complications appear to be remedied by the inclusion of the switch permutation layer. 

Although the switchless configuration manages to produce reasonable images, we conjecture that this is because the model is ``reasonably deep'' at this length scale. We suspect the predictive abilities will quickly deteriorate as $L \to \infty$ since the depth of the network only scales linearly as $\Theta(L)$.

\subsubsection{Out-of-Distribution Generalization}
\begin{figure}[htp!]
     \centering
     \includegraphics[width=0.99\textwidth,trim={00mm 0mm 00mm 00mm},clip]{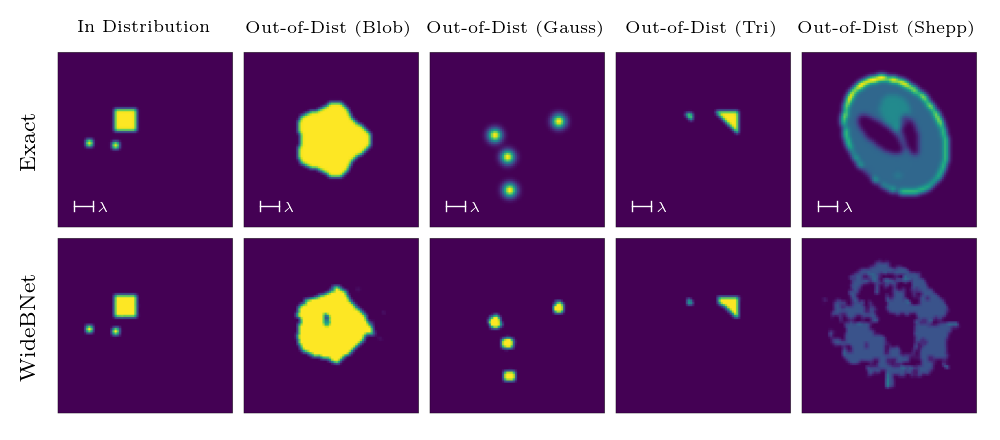}
     \caption{Out of distribution performance of a \wbnn model trained on a Square 3/5/10 dataset (example datapoint shown in left-most column). The second to fifth columns depict examples of out-of-distribution datapoints. All colour scales are normalized with respect to the exact solutions shown in the first row.  }
     \label{fig:generalization}
 \end{figure}

\rev{
We consider the performance of \wbnn on scatterers that are \emph{out-of-distribution} and distinct from the \emph{within-distribution} scatterers of the training set. The result of this experiment is shown in Fig.~\ref{fig:generalization} for a \wbnn model that is trained with randomly located squares of sidelengths 3, 5, and 10 pixels. An example datapoint from this training class is presented in the left-most column. This trained network is applied to four different scattering configurations: a non-convex shaped `blob', shown in the top row of the second column; Gaussians with characteristic length of 2 pixels, shown in the third column; triangles with sidelengths of 3 and 10 pixels, shown in the fourth column; and a Shepp-Logan phantom, shown in the last column. All colour scales are normalized with respect to the first row, which corresponds to the exact solution. 
}

\rev{
The second row of the figure depicts the output of \wbnn from noiseless, wideband, recordings of the respective scattered wavefields.  We observe that our network generalizes to two distinct, forward scattering, regimes. First, \wbnn is able to localize both the Gaussian and the triangular scatterers, in addition to resolving their shapes and sub-wavelength features. This suggests that our network learns an inverse scattering map applicable to general configurations that are dominated by Born single scattering, with seemingly no limitations on resolution. Second, we note \wbnn is also capable of generalizing to data involving strong multiple scattering, as indicated by its performance on the multiple wavelength spanning and non-convex `blob'. As Fig.~\ref{fig:generalization} demonstrates, \wbnn infills the shape, although with noticeable errors along the support of the scatter. Since during training it is only provided with data with at most three square scatterers, this infilling property suggests that our model captures some generalized properties of inverse wave scattering. We note, however, there are limits to its extrapolative ability, as suggested by the results involving the Shepp-Logan phantom. Characterizing \emph{a priori} which configurations are amenable to extrapolation remains an open problem.}

\rev{
Remarkably, other examples of neural networks extrapolating beyond the scattering configurations of their training sets have been reported in the literature. In \cite{pang2020machine} the authors apply an LSTM network, designed to explicitly incorporate the Lippmann-Schwinger kernel, to learn the physical model which generates the wavefield from scatterers. They report the ability of their network to simulate wavefields from scattering shapes unseen in training. In a similar vein, \cite{kothari2020fio} consider the inverse scattering problem, with a network architecture called FIONet which also leverages principles from Fourier integral operators, and successfully image out-of-distribution scatterers. We leave an investigation into this commonly observed extrapolation phenomena to future work.
}

\subsubsection{Partitioning of frequencies}
Table~\ref{tab:freq_ptn} reports on the difference between two competing frequency partitioning strategies: ``AllFreq'', in which the data from the entire bandwidth are fed into \wbnn at level $L$, versus ``MultiFreq'' wherein the data are only processed at the appropriate length scale $\ell$. Qualitatively both strategies produce comparable images that are sharp and resolve the sub-wavelength features. In fact, quantitatively the ``AllFreq'' strategy produces marginally lower losses (though within the same order of magnitude). However, as noted in Table~\ref{tab:freq_ptn} that the degrees of freedom of ``AllFreq'' far exceed that of ``MultiFreq''; although both strategies have the same asymptotic storage complexity of $\mathcal{O}(N\log^3(N))$ (see Section \ref{sec:network_cost}), practically speaking the constant differs by a substantial amount in favour of ``MultiFreq``.

We report that we were unable to successfully train a model by mimicking \eqref{eq:imaging_cond} directly, i.e. training single channel \wbnn models for each frequency independently, then merging their predictions via a CNN module. This is perhaps unsurprising since it is known that super-resolution algorithms require non-linear synthesis of multi-frequency data to succeed. Whereas in both ``MultiFreq'' and ``AllFreq'' this is achieved by the switch-resnet module, in this naive strategy the synthesis is performed only at the end by the CNN layers. In comparison to the optimal storage complexity of $\mathcal{O}(N\log^2 N)$ in this na\"ive strategy, note that mildly overparametrizing by a small logarithmic factor provides significant training stability to the inverse problem.

\begin{table}[t!]
\centering
\caption{Effect of frequency partition}
\label{tab:freq_ptn}
\begin{tabular}{@{}llllcll@{}} \toprule
& & \multicolumn{2}{c}{Pixel-wise Squared Loss} & \phantom{asdf} & \multicolumn{2}{c}{Image-wise Relative Loss} \\
\cmidrule{3-4}\cmidrule{6-7}
              & \textsc{DOF} & Train & Test && Train & Test \\
			  \midrule
AllFreq   & 2746368 & 2.92E-06 & 4.81E-06 &  & 1.26E-05 & 1.72E-05 \\
MultiFreq & 1913856 & 4.06E-06 & 6.40E-06 &  & 1.72E-05 & 2.27E-05\\
\bottomrule
\end{tabular}
\end{table}

\subsubsection{Training Curves \& Hyper-Parameter Sensitivity}
\label{sec:training_curves}

\subsubsection*{Training Curves} 

\begin{figure}[htp!]
     \centering
     \includegraphics[width=\textwidth,trim={00mm 0mm 00mm 00mm},clip]{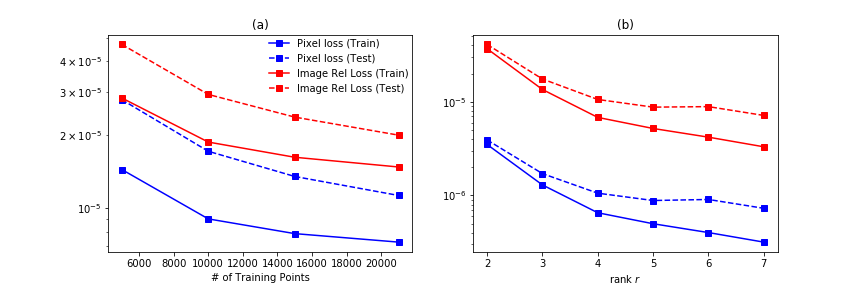}
     \caption{Sensitivity to hyper-parameters: number of training points and the rank $r$. (a) Performance of \wbnn with increasing number of training points. Note the same testing dataset, consisting of $3000$ points, was used for all experiments. Observe that the generalization gap (the distance between the dashed and solid lines) remains asymptotically as both training and testing errors saturate. This exhibits that we are saturating the model capacity. (b) Performance of \wbnn with increasing rank $r$. The testing dataset was fixed for all experiments. Note that for leaf size $s=5$ the maximum rank of the linearized model is $r_\text{max} = 25$. Although the training error decreases with increasing rank, we observe that the testing error begins to plateau beyond $r=3$.}
     \label{fig:ntrain_rank}
 \end{figure}

Figure~\ref{fig:ntrain_rank}(a) reports the training errors for models trained on datasets containing $5000$, $10000$, $15000$, and $21000$ datapoints. The trained models were evaluated on a \textit{fixed} testing set of $3000$ points (i.e. the same testing set is to compare all experiments). All remaining hyperparameters such as the learning rate and number of epochs were held the same as discussed in the beginning of Section~\ref{sec:numerical_results}. Note in all cases we remain in the over-parametrized regime since the number of datapoints is far fewer than the number of degree of freedom. Nevertheless, with only a few samples \wbnn stably achieves a pixel-wise loss on the order of $10^{-5}$.

We observe in Figure~\ref{fig:ntrain_rank} that both training and testing errors decrease with increasing training points, as expected. However, these training/testing curves quickly saturate and the differences fall less than an order of magnitude. Furthermore, the empirical generalization gap, taken to be the difference between the testing curve (dashed lines) and the training curve (solid line) remains within the same order of magnitude as the number of points is increased. These points demonstrate that our model \textit{(i)} generalizes with relatively scant training points, and \textit{(ii)} saturates its model capacity quickly, which is an indication that the architecture is well adapted to the task.

\subsubsection*{Sensitivity to the rank $r$}
While the data essentially specify the architecture through requirements on the level $L$ and leaf size $s$, it remains up to the user to select the rank $r$. We reiterate that this choice serves as a significant departure from the numerical analysis perspective of the Butterfly factorization; whereas in the original context it is essential to have the scaling $r \ll s^2$ for the purpose of fast matrix-vector multiplication\footnote{Typically the rank is determined by computations of SVDs so that $\epsilon$ is close to be machine zero. Analytical relations between $\epsilon$ and $r$ are kernel dependent and is known explicitly only in few cases.}, in the current machine learning context there is no restriction against choosing $r \geq s^2$. Nevertheless, as Figure~\ref{fig:ntrain_rank}b demonstrates, a large over-parameterization with respect to $r$ is unnecessary. Indeed, while the training metrics monotonically decrease as the model capacity increases with rank, we observe that testing errors remain relatively saturated. This suggests that performance of \wbnn~is largely insensitive to the rank and the network topology plays a more significant role.

Moreover, these results indicate that allowing for a non-uniform rank for each patch may not yield be a fruitful exercise. Or, conversely, if the intent is to compress the model further to e.g. fit on mobile devices \cite{blalock2021state}, this also suggests that tenable strategy may be to prune a trained model by adaptive patch-wise rank reduction. We leave this to future work.

\subsubsection*{Effect of CNN and ResNet Layers}
\begin{figure}[htp!]
     \centering
     \includegraphics[width=\textwidth,trim={00mm 0mm 00mm 00mm},clip]{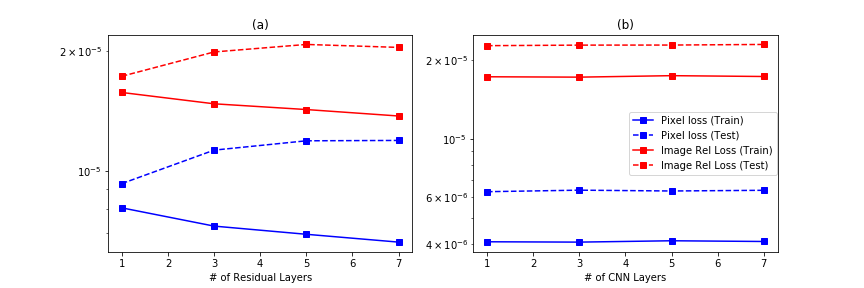}
     \caption{Sensitivity to hyper-parameters: number of residual layers and number of convolution post-processing layers. The testing set of $3000$ points was fixed across all experiments. (a) The training error decreases with increasing residual layers, but the testing error increases. Note however the variation is negligible. (These experiments all had three convolution layers) . (b) \wbnn exhibits nearly complete insensitivity to the number of $CNN$ post-processing layers. (this experiment was with three residual layers)}
     \label{fig:nCNN_nresnet}
 \end{figure}

Beyond the selection of the rank~$r$ the only remaining hyper-parameters that determine the \wbnn architecture are the number of CNN layers $N_\textrm{CNN}$ and the number of residual layers $N_\textrm{RNN}$ in the switch module. Figure~\ref{fig:nCNN_nresnet} reports on the sensitivity of the \wbnn model to these parameters. Evidently from Figure~\ref{fig:nCNN_nresnet}b we conclude that the predictive performance is unaffected by the number of post-processing CNN layers. A similar conclusion can be drawn about the number of residual layers from Figure~\ref{fig:nCNN_nresnet}a; note the fluctuations in the training and testing curves are negligible in magnitude.

\begin{table}[t!]
\centering
\caption{Training and testing errors for various datasets. Each experiment consisted of 21000 training points and \wbnn was evaluated against an independent testing dataset with 3000 points. The data were generated using a homogeneous background wavefield $c_0=1$ and data were sampled at 2.5, 5, and 10 Hz (i.e. the effective wavelength was $8$~PPW). Each row denotes a separate experiment with the scatterers consisting of triangles ($\triangle$), squares ($\square$), and \rev{Gaussian}s ($\bigcirc$). The numbers in the parentheses indicate the characteristic lengths of each scatterer; multiple numbers indicate a heterogeneous dataset of scatterer sizes, while (rot,5) indicates a dataset with rotated scatterers. In general, we observe that \wbnn does not over-fit the data. Surprisingly, on average the testing pixel-wise error decreases with decreasing length-scale.}
\label{tab:numerical_results}
\begin{tabular}{lllllll}\toprule
                          &  & \multicolumn{2}{c}{Pixel-wise Squared Loss}          &  & \multicolumn{2}{c}{Image-wise Relative Loss}           \\
\cmidrule{3-4}\cmidrule{6-7}  
Scatterer            &  & Train                  & Test     &  & Train                  & Test     \\
\midrule
$\triangle$~(3,5,10) &  & 4.06E-06               & 6.40E-06 &  & 5.38E-04               & 7.12E-04 \\
$\square$~(3,5,10)   &  & 7.12E-04               & 1.13E-05 &  & 4.63E-04               & 6.24E-04 \\
$\bigcirc$~(3,5,10)  &  & 1.24E-06               & 2.01E-06 &  & 1.89E-05               & 2.71E-05 \\
$\triangle$~(rot,5)  &  & 3.03E-06               & 4.09E-06 &  & 5.52E-04               & 7.32E-04 \\
                          &  &                        &          &  &                        &          \\
$\triangle$~(10)     &  & 2.47E-06               & 2.51E-05 &  & 9.26E-05               & 8.17E-04 \\
$\triangle$~(5)      &  & 1.14E-06               & 7.19E-06 &  & 2.11E-04               & 1.24E-03 \\
$\triangle$~(3)      &  & 4.35E-06               & 4.23E-06 &  & 2.62E-03               & 2.62E-03 \\
                          &  &                        &          &  &                        &          \\
$\square$~(10)       &  & 2.63E-06               & 7.92E-05 &  & 4.90E-05               & 1.24E-03 \\
$\square$~(5)        &  & 1.24E-06               & 2.09E-05 &  & 1.13E-04               & 1.75E-03 \\
$\square$~(3)        &  & 1.19E-05               & 1.19E-05 &  & 3.77E-03               & 3.80E-03 \\
                          &  &                        &          &  &                        &          \\
$\bigcirc$~(3)       &  & 9.89E-08               & 2.61E-06 &  & 5.97E-06               & 1.30E-04 \\
$\bigcirc$~(2)       &  & 3.19E-07               & 4.84E-07 &  & 4.35E-05               & 6.28E-05 \\
$\bigcirc$~(1)       &  & 5.86E-07               & 7.52E-07 &  & 4.71E-04               & 5.87E-04 \\
\bottomrule
\end{tabular}
\end{table}

\subsection{Heterogeneous Background}


In this section we present numerical results with scattering data from a known \emph{inhomogeneous} background medium. The variations in the background wavespeed introduce significant complications to the inverse problem. For instance, homogeneous backgrounds afford symmetries such as rotational equivariance which can be exploited for efficient network design, see e.g. \cite{FanYing:traveltime}; in an inhomogeneous background this assumption is no longer valid. The physics of wave propagation through inhomogeneous media also complicates the signal processing problem as it gives rise to multi-pathing as well as multiple arrivals due to interior scattering. While the architecture and data formatting remain unchanged, the complexity of the inverse problem for localizing scatterers, let alone super-resolution, increases in this setting. 

\begin{figure}[htp!]
     \centering
     \includegraphics[width=\textwidth,trim={00mm 00mm 00mm 00mm},clip]{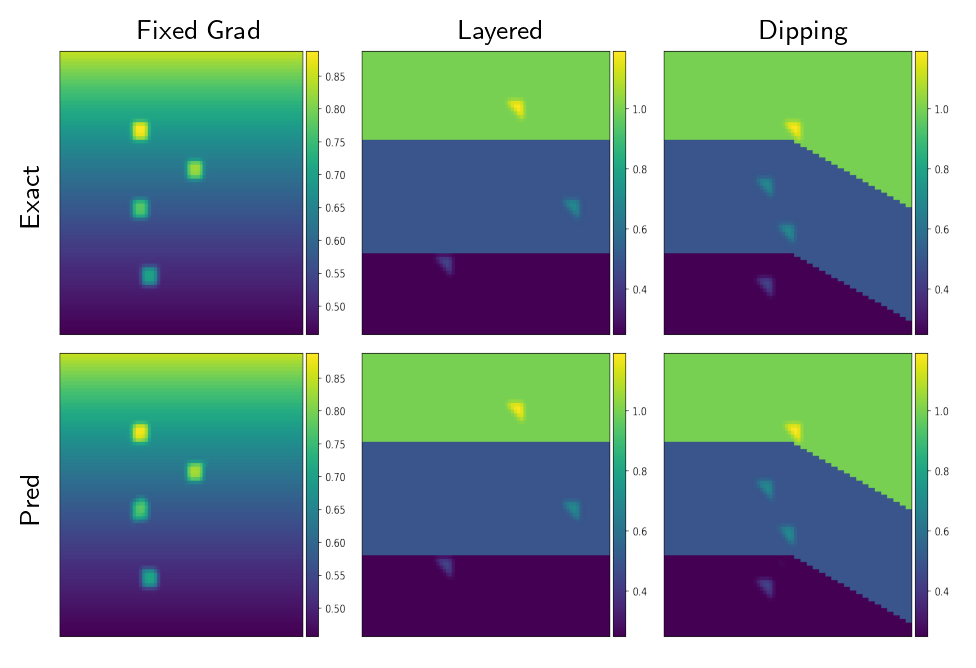}
     \caption{Visualization of \wbnn predictions on a testing set with inhomogeneous backgrounds. The colour scales of each row is normalized to the first column. The background medium is assumed known. (a) With a linearly increasing gradient in the background. (b) A layered medium increasing velocity with depth. (c) A layered medium with dipping reflectors. Note that the last configuration is not translation invariant.}
     \label{fig:inhomogeneous}
 \end{figure}

We tested the algorithm for two heterogeneous backgrounds: (i) a smooth linearly increasing background medium with wavespeed $c=0.5$ at the top and $c=1.5$ at the bottom, and (ii) layered background medium with wavespeeds $c=1$, $c=2$, and $c=4$. The results of trained \wbnn models on testing data are shown in Figure~\ref{fig:inhomogeneous}. We observe in Figure~\ref{fig:inhomogeneous}(b) that \wbnn manages to process the multiple arrivals to image the triangular scatterers. However, surprisingly, it does significantly poorer for the smoothly varying background. Explaining this discrepancy remains an open problem. 

\textbf{Remark:} The notion of resolution becomes ambiguous for inhomogeneous media as the wavelength changes with background medium $c(x,z)$ following the dispersion relation in \eqref{eq:dispersion}. Nevertheless, across the range of background velocities the scatterers still contain sub-wavelength features such as e.g. the corners.

\subsubsection{Comparison versus FWI}

\rev{We compare \wbnn against FWI, implemented in Matlab, inverting for the same perturbation. The descent path is initialized with the known homogeneous background, and the gradient is computed using standard adjoint state methods. We selected as the optimization method L-BFGS implemented using the \texttt{fminunc} routine.}

\rev{Following standard practices in the geophysical community we use a frequency sweep to regularize against the non-convexity of the objective function. We tested a dozen frequency combinations, and we selected the one which produced the best images. In the sweep, the data at different frequencies are fed to the optimization loop at three stages. At each stage we process data only at a certain frequency, without combining them, but we use the estimate at the end of one stage to initialize the subsequent stage one: in the first stage we process the lowest frequency data, we save the final answer which will be used as an initial guess for the next stage, which will process data in the immediately higher frequency-band, and we repeat until  data at all frequencies are processed.}

\begin{figure}[htp!]
     \centering
     \includegraphics[width=0.98\textwidth,trim={00mm 0mm 00mm 00mm},clip]{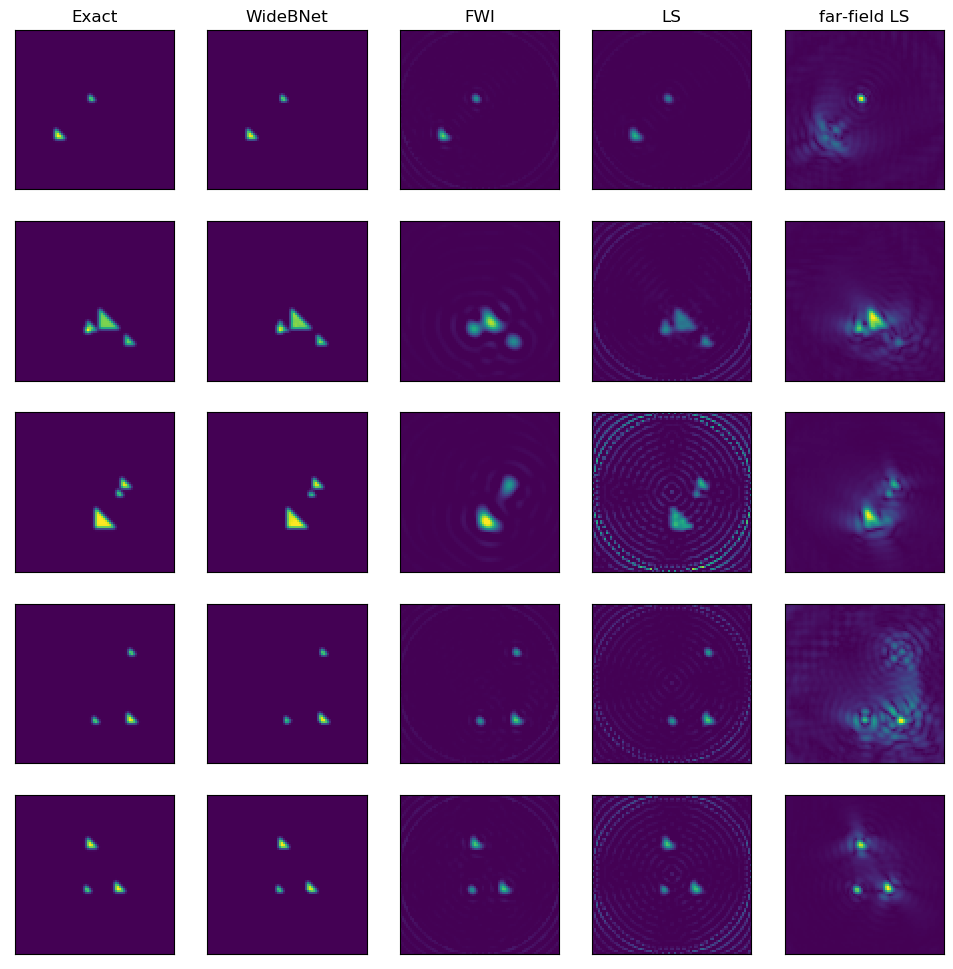}
     \caption{From left to right columns: exact perturbation, prediction by \wbnn, reconstruction by FWI using data at $\{2.5Hz, 5.0Hz, 10.0Hz\}$, least-squares using the finite difference modeling and data at $\{2.5\textrm{Hz}, 5.0\textrm{Hz}, 10.0\textrm{Hz}\}$, least-square using far-field approximation and data at $\{10.0Hz\}$}
     \label{fig:comparison_fwi}
 \end{figure}

\rev{We ran the optimization until either the residual stagnated around $10^{-6}$, or the norm of the gradient fell below $10^{-4}$. In order to avoid the inverse crime we use a fourth order finite different stencil in the FWI formulation, in contrast to the data which was generated using a five-point second order stencil. For completeness, we also computed the regularized least-squares (LS) estimate using the far-field asymptotics in \eqref{eq:far_field_pattern}, using only the highest frequency data, and the least squares (LS) estimate using the finite difference discretization of the problem (with $9$-point stencil) with wide-band data. After a laborious search for the best reconstruction we found that regularization parameter $\epsilon = 1$ for LS produced the best localisation while simultaneously minimizing oscillatory artifacts. The linear system was solved using \texttt{gmres} with a tolerance of $10^{-3}$. In Figure~\ref{fig:comparison_fwi} we can observe that for this specific class of scatters \wbnn outperforms all the other methods, and provides a sharper image of the perturbation with the correct amplitude (the far-field LS was re-scaled in this case).}

\rev{We can observe that the reflectors are properly placed but the result from our neural network provides a better localization, sharper corners, with far fewer oscillatory artifacts. We point out that procuring these images for FWI was labour- and time-intensive. It took roughly a day to test all the different frequency sweeps and the full computation. The full computation took roughly one minute and a half in average for FWI, around two minutes for LS and half a minute for far-field LS. The experiments were carried in a $16$-core workstation with an AMD 2950X CPU and $128$ GB of RAM. In contrast, the training stage for \wbnn took in total 12 hours, and the inference takes a fraction of a second, running on an Nvidia GTX 1080Ti graphics card. }


\section{Conclusion \& Future Work}

In this manuscript we have designed an end-to-end architecture that is specifically tailored for solving the inverse scattering problem. We have shown that by assimilating multi-frequency data and coupling them through non-linearities  we can produce images that solve the inverse scattering problem. Our tool produces results which are competitive with optimization-based approaches, but at a fraction of the cost. More critically, we have demonstrated that our architecture design and data assimilation strategy avoids three known shortcomings with conventional architectures and also other butterfly-based networks: (i) by incorporating tools from computational harmonic analysis, such as the butterfly factorization, and multi-scale methods, such as the Cooley-Tukey FFT algorithm, we are able to drastically reduce the number of trainable parameters to match the inherent complexity of the problem and lower the training data requirements, (ii) our network has stable training dynamics and does not encounter issues such poorly conditioned gradients or poor local minima, and (iii) our network can be initialized using standard off-the-shelf technologies.

In addition, we have shown that our network recovers features below the diffraction limit of general, albeit fixed, class of scatterers.  Even though there is an underlying assumption on the distribution of the scatterers we do not explicitly exploit it. Thus, one future research direction is to use the current architecture within a VAE or GAN framework, to fully capture the underlying distribution, and to further study the limits of the current architecture to image sub-wavelength features. Following the same approach one can seek to extend the applicability of the current architecture to the cases where there is noise in signal, or uncertainty on the background medium. 


\section*{Acknowledgments}
We thank Yuehaw Khoo, Lexing Ying, Guillaume Bal, Yingzhou Li, Zhilong Fang, Pawan Bhawadraj, and Nori Nakata for fruitful discussions. We also thank George Barbastathis for detailed feedback on an earlier draft, and for invaluable references. In addition, we thank the two anonymous referees for their helpful comments and suggestions.

\bibliographystyle{siamplain}
\bibliography{references_widebnet.bib}

\newpage
\appendix

\section{Permutation and Switch Indices}
\label{sec:perm_switch_idx}

The permutation indexing that enables the $G^\ell$ and $H^\ell$ layers to be described as block-diagonal matrices can be derived from an involved analysis of the two-dimensional butterfly factorization. For convenience, we provide a generic pseudocode in Alg.~\ref{lst:permidx} which automates this construction for all choices of $L$. We note that this permutation assumes the input vector is Morton-flattened. Similarly, Alg.~\ref{lst:switchidx} yields pseudocode for generating the specific switch permutation indices of the Butterfly algorithm.

\begin{lstlisting}[mathescape=true,
floatplacement=tb,label={lst:permidx},
caption={Pseudo code for permutation pattern $\pi_\ell$ used in layers $G^\ell$ and $H^\ell$ when processing Morton-flattened inputs.}
]
def permutation_indices($L$, $\ell$):
  # indices inside each $4^{L-\ell}\times 4^{L-\ell}$ block
  $\Delta = 4^{L-\ell-1}$

  # [0,$\Delta$,$2\Delta$,$3\Delta$,0,$\Delta$,$2\Delta$,$3\Delta$,$4\Delta$,\ldots]
  $\pi^\ell$ = np.file(np.arange(4)*$\Delta$, $\Delta$)

  # + [0,0,0,0,1,1,1,1,\ldots,$\Delta$, $\Delta$,$\Delta$,$\Delta$]
  $\pi^\ell$ += np.repeat(np.arange($\Delta$), 4)

  # indices for entire block diagonal matrix
  $\pi^\ell$ = np.tile($\pi^\ell$, $4^\ell$)
  $\pi^\ell$ += np.repeat(np.arange($4^\ell$)*$4^{L-\ell}$, $4^{L-\ell}$)

  return $\pi^\ell$


\end{lstlisting}

\begin{lstlisting}[mathescape=true,
floatplacement=tb,label={lst:switchidx},
caption={Pseudo code for switch permutation indices for processing Morton-flattened inputs.}
]
def switch_indices($L$):
  $\pi_\textrm{switch}$ = np.arange($2^L$)*($2^L$)

  $\pi_\textrm{switch}$ = np.tile($\pi_\textrm{switch}$, $2^L$)
  $\pi_\textrm{switch}$ += np.repeat(np.arange($2^L$), $2^L$)

  return $\pi_\textrm{switch}$
\end{lstlisting}

%
%

\end{document}